\documentclass[10pt,journal,compsoc]{IEEEtran}
\usepackage{amsmath,amsfonts}
\usepackage{algorithmic}
\usepackage{algorithm}
\usepackage{array}
\usepackage[caption=false,font=normalsize,labelfont=sf,textfont=sf]{subfig}
\usepackage{textcomp}
\usepackage{stfloats}
\usepackage{url}
\usepackage{verbatim}
\usepackage{graphicx}
\usepackage{cite}
\usepackage{xcolor}
\usepackage{colortbl} 
\usepackage{enumitem}
\usepackage{multirow}
\usepackage{float}
\usepackage{booktabs}
\usepackage{setspace}

\begin{document}

\title{Exploring Context Generalizability in Citywide Crowd Mobility Prediction: An Analytic Framework and Benchmark}

\author{Liyue Chen, Xiaoxiang Wang, Leye Wang
\IEEEcompsocitemizethanks{\IEEEcompsocthanksitem Liyue Chen and Leye Wang are with the Key Lab of High Confidence Software Technologies, Ministry of Education, China, and School of Computer Science, Peking University, Beijing 100871, China. E-mail: chenliyue2019@gmail.com, leyewang@pku.edu.cn.\protect\\
\IEEEcompsocthanksitem Xiaoxiang Wang is with the School of Software \& Microelectronics, Peking University, Beijing 102600, China.\protect\\
}
\thanks{This work was supported by National Key Research and Development Program of China (2023YFB3308504). (Corresponding author: Leye Wang.)}
}

\markboth{Journal of \LaTeX\ Class Files,~Vol.~14, No.~8, August~2021}%
{Shell \MakeLowercase{\textit{et al.}}: A Sample Article Using IEEEtran.cls for IEEE Journals}

\IEEEpubid{0000--0000/00\$00.00~\copyright~2021 IEEE}

\IEEEtitleabstractindextext{
\begin{abstract}
Contextual features are important data sources for building citywide crowd mobility prediction models. However, the difficulty of applying context lies in the unknown generalizability of contextual features (e.g., weather, holiday, and points of interests) and context modeling techniques across different scenarios. In this paper, we present a unified analytic framework and a large-scale benchmark for evaluating context generalizability. The benchmark includes crowd mobility data, contextual data, and advanced prediction models. We conduct comprehensive experiments in several crowd mobility prediction tasks such as bike flow, metro passenger flow, and electric vehicle charging demand. Our results reveal several important observations: (1) Using more contextual features may not always result in better prediction with existing context modeling techniques; in particular, the combination of holiday and temporal position can provide more generalizable beneficial information than other contextual feature combinations. (2) In context modeling techniques, using a gated unit to incorporate raw contextual features into the deep prediction model has good generalizability. Besides, we offer several suggestions about incorporating contextual factors for building crowd mobility prediction applications. From our findings, we call for future research efforts devoted to developing new context modeling solutions.
\end{abstract}

}
\maketitle

\IEEEdisplaynontitleabstractindextext
\IEEEpeerreviewmaketitle

\section{Introduction}
\IEEEPARstart{U}nderstanding and modeling citywide crowd mobility is of significant importance for the advancement of smart city applications, encompassing critical fields such as safety and security \cite{ridesharing_car_detection_2018,traffic_connected_2022}, crowd sensing \cite{task_allocation_2023}, personalized service recommendation \cite{systematic_2022,cell_traj_prediction_2022}, resource allocation and energy saving \cite{c_ran_mobility_2021, step_2021, GASTN_2022, redpacketbike_2022, GTCN_2022, MVSTGN_2023}, transportation planning \cite{mobility_driven_2023} and urban infrastructure development \cite{city_transfer_2018}. Accurate prediction of citywide crowd mobility is a fundamental and essential research challenge, representing a crucial abstraction of numerous tasks including bike-sharing flow forecasting \cite{li_traffic_2015,yang_mobility_2016,chai_multi_graph_2018,LiBikeTKDE2019}, ride-sharing demand predicting \cite{tong_simpler_2017,ke_short-term_2017,wang_deepsd:_2017,yao2018DeepMulti,Saadallah_taxi_2020,one4all_icde_2024}, and traffic speed prediction \cite{li2017diffusion,yi2019citytraffic}. These predictive capabilities facilitate the optimization and effective implementation of various urban services, enabling informed decision-making and resource allocation in a rapidly evolving city environment. 

Contextual factors (e.g., weather, holidays) have proven beneficial for citywide crowd mobility prediction \cite{zhang2017deep,demand_cellular_2022}. Taking weather context as an example, rising temperatures will promote the usage of bike-sharing \cite{li_traffic_2015}, and heavy rains will lessen the usage of both bike-sharing and online ride-hailing \cite{hoang_fccf:_2016}. In general, context is any information that can be used to characterize the situation of an entity, where an entity can be a person, place, or physical/computational object \cite{abowd1999towards}. Comprehending and modeling contextual factors assumes critical significance in crowd mobility prediction tasks, as it has the potential to furnish valuable supplementary insights into the acquisition of specific spatiotemporal patterns. By incorporating context awareness into predictive models, an enriched understanding of crowd dynamics can be achieved, leading to more accurate and robust mobility forecasts \cite{xu2016context, zhang2017deep, lin2019deepstn+}.

To achieve accurate crowd mobility prediction, researchers have recognized the importance of considering spatial and temporal correlations and have proposed various techniques to model spatiotemporal dependencies, including attention mechanisms \cite{liang_geoman:_2018, ASTGCN_2019, GMAN_AAAI2020} and graph convolution networks \cite{li2017diffusion,yu2018spatio,chai_multi_graph_2018}. This problem is commonly referred to as spatiotemporal crowd flow prediction (STCFP) \cite{zhang2017deep, regiontrans_2019, flexible_partition_2020} or spatiotemporal traffic prediction (STTP) problem \cite{buildsen_2021, STMeta}. Temporal correlations elucidate the connection between future and past flows, whereas spatial correlations primarily capture the interrelationship between different geographical locations. Furthermore, extensive studies have been conducted to explore the generalizability of spatiotemporal correlations \cite{STMeta}. By comprehending and leveraging these spatiotemporal correlations, the predictive capabilities of spatiotemporal models can be enhanced, leading to more robust and reliable predictions of crowd mobility.

However, despite the utilization of specific contextual features in certain applications \cite{li_traffic_2015, deep_fusion_net_2018}, the generalizability of contextual features is still under-investigated. For example, POIs (Points-Of-Interests) data is effective in taxi demand prediction problems \cite{tong_simpler_2017}, but whether POIs data is still beneficial in other scenarios remains to be evaluated. Besides, while pioneering studies propose various context modeling techniques such as \emph{Adding} \cite{zhang2016dnn}, \emph{Embedding} \cite{chai_multi_graph_2018},  and \emph{Gating} \cite{zhang_flow_2019}, it is still hard to select appropriate context modeling techniques for a given problem since the generalizability of these modeling techniques is unknown.

Overall, analyzing the generalizability of contextual features and context modeling techniques is of significant value for building effective mobility prediction models. Meanwhile, this analysis is challenging in the following aspects:

\textbf{\textit{Unknown generalizability of contextual features}}. Most studies directly consider a specific set of contextual features without carefully analyzing whether the selection of these features is optimal. To the best of our knowledge, no previous study has thoroughly compared the generalizability of contextual features across different application scenarios. 

\textbf{\textit{Lacking a taxonomy of context modeling techniques}}. Previous studies propose various context modeling techniques. For the convenience of analyzing the generalizability of context modeling techniques, a comprehensive taxonomy is desired.

\textbf{\textit{Unknown generalizability of context modeling techniques}}. Though several context modeling techniques have been proposed, it is hard to determine a suitable modeling technique given an STCFP task. As far as we know, no previous study has analyzed the generalizability of different context modeling techniques across scenarios.

To sum up, in this paper, we aim to analyze the generalizability of both contextual features and context modeling techniques in STCFP applications. We conduct a large-scale analytical and experimental empirical study. Particularly, we try to give some design guidelines for the research community and make the following contributions:
\begin{itemize}

\item To the best of our knowledge, this is the first study that focuses on investigating how contextual factors (both features and modeling techniques) would generally and quantitatively impact crowd mobility prediction performance in various practical scenarios. By answering this critical but under-investigated problem, we expect that this research can inspire researchers and practitioners to learn how to efficiently incorporate contextual factors in crowd mobility prediction models and applications.
	
\item By surveying recent studies, we categorize the mostly-used contextual features in literature, including weather, holiday, temporal position, POIs, road, demographic, and spatial position. Additionally, a new unified analytic framework for context modeling techniques in STCFP models is developed, summarizing a total of 14 techniques. While ten techniques are leveraged in literature, \textit{the other four are newly established and investigated by our work}.

\item We have created and released an experimental repository\footnote{\url{https://github.com/Liyue-Chen/STCFPContext}} that provides crowd mobility data, aligned contextual data, and state-of-the-art STCFP models. The repository covers three applications: bike-sharing usage, metro passenger flow, and electronic vehicle demand. Researchers can utilize this platform to experiment with their own STCFP applications and explore the effectiveness and generalizability of the context.

\item Our extensive experiments reveal a surprising and counter-intuitive finding --- with current context modeling techniques, \textit{introducing more contextual features (e.g., weather) would often degrade the prediction performance}. However, we find that holiday and temporal position features generally provide beneficial information, while weather and POIs do not offer significant improvements and may even have negative impacts. Based on these observations, we provide suggestions for crowd mobility prediction models to incorporate contextual factors. Furthermore, we emphasize the need for future research to develop new context processing and modeling solutions that fully exploit the potential of contextual features.
\end{itemize}

\section{Problem and Generalizability}
\subsection{Problem Formulation}
\textbf{\textit{Definition 1 (Location)}.} 
The set of $N$ locations where the crowd flow are observed is denoted as $\mathcal{L}=\{l_1,l_2,...,l_N\}$. The locations can both represent the equal-sized grid in dock-less applications (e.g., ride-sharing \cite{tong_simpler_2017,ke_short-term_2017,wang_deepsd:_2017,zhu_deep_2017,yao2018DeepMulti,Saadallah_taxi_2020}) or the dock stations in dock-based applications (e.g., metro passenger flow \cite{liu_deeppf:_2019}).

\noindent \textbf{\textit{Definition 2 (Crowd flow time series)}.} Given a set of $N$ locations $\mathcal{L}$, $\textbf{X}_t \in \mathbb{R}^{N\times D}$ is the crowd flow observations of $N$ locations during the time interval $t$. $D$ is the feature dimension of a location. The crowd flow time series from time interval 1 to the time interval $t$ is $\mathcal{X}_{1:t} = \{\textbf{X}_1,....,\textbf{X}_t\}$

\noindent \textbf{\textit{Definition 3 (Temporal contextual features)}.} 
Temporal contextual features are characterized by their variabilities within short time periods, such as one hour or one day. One example of such features is the weather, which can change abruptly from sunny to rainy within a short time interval. In urban applications, these features tend to be consistent across different locations at the same moment. For brevity, we will refer to these features as temporal contextual features, which are primarily sensitive to changes in the temporal dimension. The temporal contextual features from time interval 1 to the time interval $T$ can be represented as $\mathcal{T}_{1:T} \in \mathbb{R}^{T \times E_t}$, where $E_t$ denotes the feature dimension of the temporal context.

\noindent \textbf{\textit{Definition 4 (Spatial contextual features)}.} 
Spatial contextual features exhibit variability in their locations, with Points of Interest (POIs) being one example. POIs in business areas are typically more numerous than those in industrial parks. These features are known to remain relatively stable over long periods (POIs records from OpenStreetMap only change a few samples in one day). Notably, spatial contextual features are primarily sensitive to changes in the spatial dimension. In this regard, we represent the spatial contextual features of $N$ locations using $\mathcal{S} \in \mathbb{R}^{N\times E_s}$, where $E_s$ refers to the feature dimension of the spatial context.

\noindent \textbf{\textit{Context-engaged Spatio-Temporal Crowd Flow Prediction Problem}.}
Suppose that we have a set of $N$ locations $\mathcal{L}$. Given $\mathcal{X}_{(t-P):t}$, the past $P$ step historical crowd flows observation, we aim to learn a function $f$ which maps $\mathcal{X}_{(t-P):t}$ and the corresponding temporal contextual features $\mathcal{T}_{(t-P):t} \in \mathbb{R}^{P\times E_t}$ and spatial contextual features $\mathcal{S} \in \mathbb{R}^{N\times E_s}$ to the observation in the next time slots $t+1$:
\begin{equation}
    [\mathcal{X}_{(t-P):t}, \mathcal{T}_{(t-P):t}, \mathcal{S}] \xrightarrow{f} \hat{\textbf{X}}_{t+1}
\end{equation}
Many real-world applications can be formulated as the above problem, such as bike-sharing \cite{li_traffic_2015,yang_mobility_2016,chai_multi_graph_2018,LiBikeTKDE2019}, ride-sharing \cite{tong_simpler_2017,ke_short-term_2017,wang_deepsd:_2017,zhu_deep_2017,yao2018DeepMulti,Saadallah_taxi_2020}, metro passenger flow \cite{liu_deeppf:_2019, wang_subways_two_way_2022, adaptive_fusion_metro_2023} and so on.

\subsection{Generalizability}
In this paper, we mainly focus on two context generalizability issues rather than specific applications: \textit{Feature Generalizability} and \textit{Technique Generalizability}:

\textbf{\textit{Feature Generalizability}}. As existing STCFP applications may involve different kinds of contextual features (e.g., weather and holiday for bike-sharing applications\cite{chai_multi_graph_2018}, weather and POIs for ride-sharing applications \cite{Saadallah_taxi_2020}. More details are in Table \ref{tab:features}), the feature generalizability refers to: \textit{Given several kinds of contextual features, which feature combination is generalizable to be effective for various applications?}

\textbf{\textit{Technique Generalizability}}: Previous researchers have proposed many context modeling techniques (e.g., \textit{gating} mechanism \cite{zhang_flow_2019, sunIrregular} and \textit{Concatenate} \cite{zhu_deep_2017,chai_multi_graph_2018,liu_deeppf:_2019,yuan_demand_2021,liang_geoman:_2018, Zhang_Huang_Xu_Xia_Dai_Bo_Zhang_Zheng_2021}) and many spatiotemporal dependencies modeling techniques (e.g., \textit{STGCN} \cite{yu2018spatio}, \textit{GraphWaveNet} \cite{graphwavenet_2019}, and \textit{AGCRN}\cite{AGCRN_2020}) for a variety of applications. The technique generalizability refers to: (1) \textit{Given a certain kind of context modeling technique, is it generalizable to be effective for various applications?} (2) \textit{Regarding different spatiotemporal modeling techniques, is a context modeling technique generalizable to be effective for various spatiotemporal models?}

\section{Analytical Studies on Contextual Features}
\begin{figure}[h]
	\centering
	\includegraphics[width=.7\linewidth]{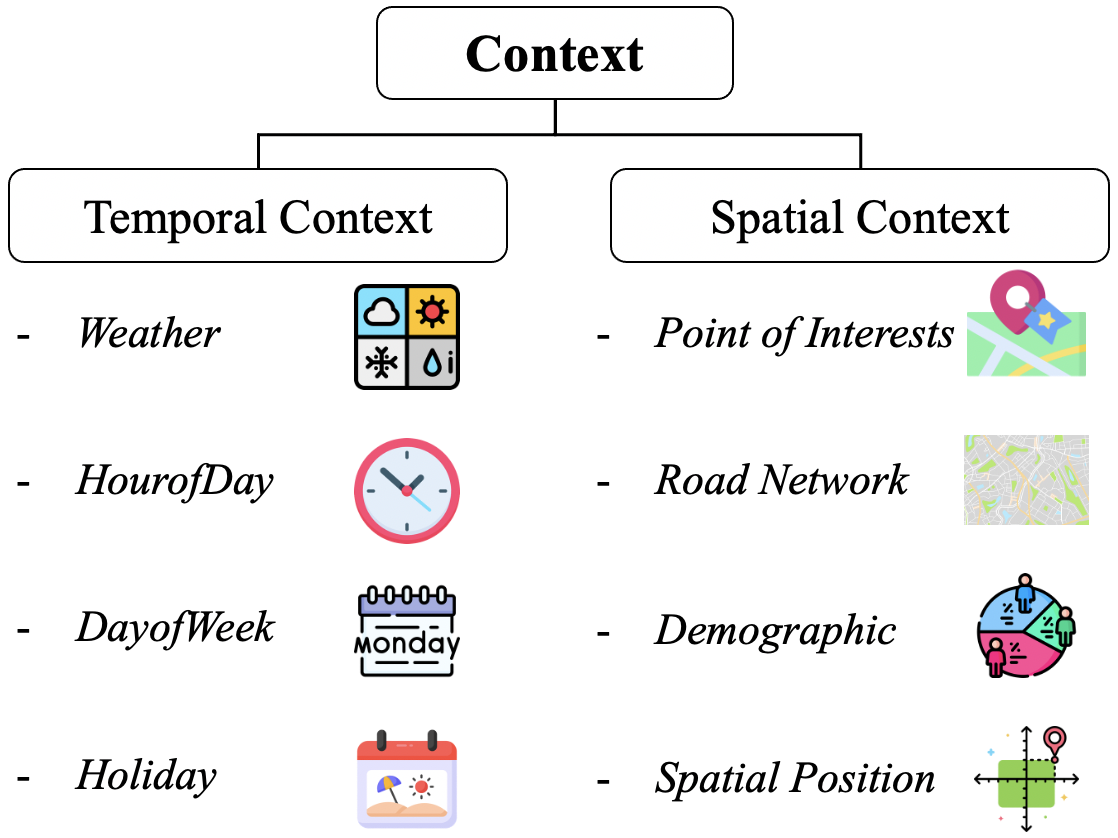}
    \vspace{-1em}
	\caption{Examples of temporal and spatial contextual features.}
	\label{fig: feature}
    \vspace{-.5em}
\end{figure}

In this section, we first introduce the methodology for investigating contextual features. We then elaborate on the selected contextual features, which are categorized into temporal and spatial contexts as shown in Figure~\ref{fig: feature}. Finally, we discuss contextual data preprocessing methods.

\subsection{Methodology for Investigating Contextual Features}
\begin{figure}[h]
	\centering
	\includegraphics[width=.95\linewidth]{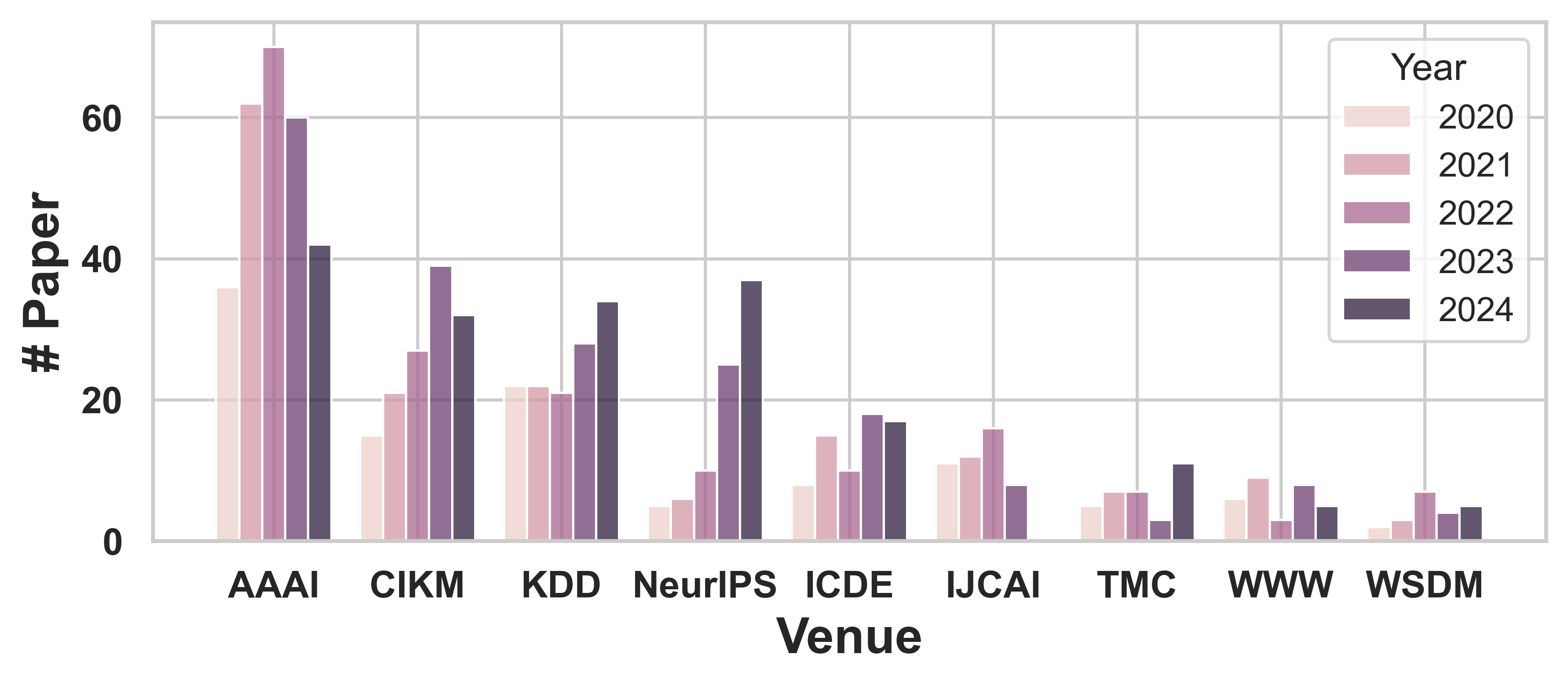}
    \vspace{-1.5em}
	\caption{Statistics on recent crowd mobility prediction papers, covering the period up to November 2024.}
	\label{fig: st_paper_statistics}
    \vspace{-.5em}
\end{figure}

To comprehensively examine the use of contextual features, we survey crowd mobility prediction papers from nine reputable venues, including IEEE TMC, KDD, ICDE, WWW, WSDM, CIKM, NeurIPS, AAAI, and IJCAI. We investigate whether each paper includes contextual features and identify the specific types of context used. Figure~\ref{fig: st_paper_statistics} shows the number of crowd mobility prediction papers published by each venue from 2020 to November 2024.\footnote{The paper list is available at \url{https://github.com/uctb/ST-Paper}.} In recent years, there have been over 100 mobility prediction papers annually, highlighting the practical significance of the spatiotemporal mobility prediction problem.

Furthermore, Table~\ref{tab:features} presents the types of contextual features utilized in these studies. In this paper, we mainly focus on publicly accessible contextual features (e.g., weather and holidays), as they may be more suitable than private data when developing prediction models for new applications. As a result, we have selected three kinds of temporal contextual features (i.e., weather, holiday, and temporal position) and four spatial contextual features (i.e., points of interests, road, demographic, and spatial position). We also believe that institutes' private data (e.g., building sensing data \cite{buildsen_2021}) also have great potential for crowd mobility prediction. However, due to their difficulty and high cost in obtaining, they are not included in this benchmark study.

\begin{table*}[ht]
	\footnotesize
	\caption{Contextual features and context modeling techniques in STCFP studies. `TP' and `SP' are temporal and spatial position features (e.g., time of day and geographical coordinates), respectively. (T: Temperature; H: Humidity; V: Visibility; WS: Wind Speed; WD: Wind Degree; AQ: Air Quality; S: Weather State)}
    \vspace{-1.5em}
	\label{tab:features}
	\begin{center}
			\begin{tabular}{lcccccccccc}
\toprule
\multirow{2}{*}{\textbf{Applications}}& \multicolumn{3}{c}{\textbf{Temporal Contextual Features}} & \multicolumn{4}{c}{\textbf{Spatial Contextual Features}} & \multirow{2}{*}{\textbf{Modeling Technique}} \\
\cmidrule(lr){2-4} \cmidrule(lr){5-8}
&\textbf{Weather} & \textbf{Holiday} & \textbf{TP} & \textbf{POIs} & \textbf{Road} & \textbf{Demographic} & \textbf{SP}  \\
\midrule
\multicolumn{2}{l}{\textbf{Bike-sharing}}\\

\emph{Li et al.} \cite{li_traffic_2015}& T;WS;S & & & & & & & \textit{Feature Concatenate} \\
\emph{Yang et al.} \cite{yang_mobility_2016}& T;H;V;WS;S & & \checkmark &  & & & & \textit{Feature Concatenate}\\
\emph{Chai et al.} \cite{chai_multi_graph_2018}& T;WS;S & \checkmark & & & & & & \emph{Emb-Concat}\\
\emph{Li et al.} \cite{LiBikeTKDE2019}& T;WS;S & \checkmark & \checkmark &  & & & &\textit{Feature Concatenate}\\
\textit{He et al.} \cite{he_escooter_2022} & T;H;WS;S & & & \checkmark & & \checkmark & & \emph{Emb-Add} \\
\textit{Jia et al.} \cite{tisc_tmc_2024} & T;WS;S & \checkmark & \checkmark  &  & &  & & \emph{Emb-Add} \\

\midrule
\multicolumn{2}{l}{\textbf{Ride-sharing}}\\
\emph{Tong et al.} \cite{tong_simpler_2017}& T;H;WS;WD;AQ;S &\checkmark &\checkmark & \checkmark & & & & \textit{Feature Concatenate}\\
\emph{Ke et al.} \cite{ke_short-term_2017}& T;H;V;WS;S & &\checkmark & & & & & \emph{LSTM-Add}\\
\emph{Wang et al.} \cite{wang_deepsd:_2017}& T;AQ;S & & \checkmark & & & & \checkmark & \emph{MultiEmb-Concat}\\
\emph{Zhu et al.} \cite{zhu_deep_2017} & & \checkmark & & & & & & \emph{Raw-Concat}\\
\emph{Yao et al.} \cite{yao2018DeepMulti}& T;S &\checkmark& & & & & & \emph{EarlyConcat}\\
\emph{Saadallah et al.} \cite{Saadallah_taxi_2020} & T;WS;S & & & \checkmark & & & &  \textit{Feature Concatenate}\\		

\midrule
\multicolumn{2}{l}{\textbf{Metro Passenger Flow}}\\
\emph{Liu et al.} \cite{liu_deeppf:_2019}& S & &\checkmark & & & & &\emph{MultiEmb-Concat}\\
\emph{Wang et al.} \cite{wang_subways_two_way_2022} & & & &\checkmark & &\checkmark & & \textit{Feature Concatenate} \\
\textit{Xu et al.} \cite{adaptive_fusion_metro_2023} & T;H;WS;V;AQ & \checkmark & \checkmark & & & & & \textit{Raw-Gating}  \\
\midrule
\textbf{Traffic Flow}\\
\emph{Yi et al.} \cite{yi2019citytraffic}& & & \checkmark& & & & & \emph{Emb-Add}\\
\emph{Barnes et al.} \cite{BarnesBCFTX20} & & & \checkmark & & & & & \textit{Feature Concatenate} \\
\emph{Zheng et al.} \cite{zheng_deepstd_2020} & T;WS;V;S & \checkmark & \checkmark & \checkmark & & & & \emph{Feature engineering + CNN} \\
\emph{Chen et al.} \cite{chen_gated_2020} & S & \checkmark & \checkmark & & & & &\emph{Emb-Concat}\\
\textit{Zheng et al} \cite{GMAN_AAAI2020} & & & \checkmark & & \checkmark  & & & \emph{EarlyConcat} + \emph{Raw-Concat}\\
\textit{Pan et al.} \cite{meta_learning_traffic_2019} & & & & \checkmark & \checkmark & & & \textit{Context-Gating} \\
\emph{Zhang et al.} \cite{Zhang_Huang_Xu_Xia_Dai_Bo_Zhang_Zheng_2021}& WS;T;S & \checkmark &\checkmark & & & & &\emph{MultiEmb-Concat}\\
\emph{Yuan et al.} \cite{yuan_demand_2021} & S & \checkmark & \checkmark& & & & & \emph{Emb-Concat} \\
\textit{Kim et al.} \cite{traffic_adversarial_2022} & T;S & & \checkmark & & & & & \textit{EarlyConcat}\\
\textit{Han et al.} \cite{han_traffic_activity_2023} & & & & & \checkmark & & \checkmark & \emph{EarlyConcat} + \emph{Raw-Concat} \\
\textit{Wang et al.} \cite{stone_kdd_2024} & & \checkmark & \checkmark & &  & & & \textit{Similarity Measurement} \\
\textit{Zhou et al.} \cite{cdsa_tmc_2024} & H;WS;S & & \checkmark &  & & & \checkmark & \textit{Emb-Concat}\\
\midrule
\textbf{Citywide Crowd Flow}\\
\emph{Hoang et al.} \cite{hoang_fccf:_2016}& T & & & & & & & \textit{Feature Concatenate} \\
\emph{Zhang et al.} \cite{zhang2016dnn}& & &\checkmark & & & & & \emph{Raw-Add}\\
\emph{Zhang et al.} \cite{zhang2017deep}& T;WS;S & \checkmark & & & & & &\emph{Emb-Add}\\
\emph{Zonoozi et al.} \cite{zonoozi2018}& & &\checkmark & & & & & \emph{Raw-Add}\\
\emph{Lin et al.} \cite{lin2019deepstn+}& &  & \checkmark & \checkmark & & & & \emph{EarlyConcat}\\
\emph{Zhang et al.} \cite{zhang_flow_2019}& T;WS;S & \checkmark & & & & & & \emph{Raw-Gating}\\
\emph{Chen et al.} \cite{wang_dynamic_nodate}& S & &\checkmark & & & & & \emph{EarlyConcat}\\
\emph{Sun et al.} \cite{sunIrregular}& T;WS;S & \checkmark & \checkmark & & & & & \emph{Emb-Gating/Add}\\
\emph{Jiang et al.} \cite{jiang_flow_2021} & & & \checkmark & & & & & \emph{EarlyConcat} \\
\emph{Liang et al.} \cite{fine_grained_2021} & T;WS;S & \checkmark & & \checkmark & \checkmark & & & \emph{EarlyConcat} \\

\textit{Wang et al.} \cite{ambulance_prediction_2021} & & & & &\checkmark & \checkmark & & Similarity Graph + GNN \\
\emph{Chen et al.} \cite{travel_time_cooperative_2022} & & \checkmark & \checkmark & & \checkmark & & & \textit{Feature Concatenate} \\
\textit{Yao et al.} \cite{MVSTGN_2023} & & \checkmark & \checkmark & \checkmark & & & & \textit{Emb-Add}  \\
\textit{Zhao et al.} \cite{semantic_urban_flow_2022} & T;WS;S & \checkmark & \checkmark & & & & & \textit{Emb-Add} \\
\textit{Jin et al.} \cite{cross_tres_2022} & & & & \checkmark & \checkmark & & & \textit{Similarity Measurement}\\
\textit{Wang et al.} \cite{CityCAN_wsdm_2024} & S & & \checkmark & & & & \checkmark & \textit{EarlyAdd} \\
\textit{Cai et al.} \cite{ctp_tmc_2024} &  & \checkmark & \checkmark &  & & &  & \textit{Emb-Concat} \\
\bottomrule
\end{tabular}
\end{center}
\vspace{-1em}
\end{table*}

\begin{table*}[ht]
	\footnotesize
	\caption{The most influential weather features and the most correlated spatial contextual features in the Bike Chicago dataset.}
    \vspace{-1em}
	\label{tab: correlation_score}
	\begin{center}
			\begin{tabular}{lcccccccccc}
\toprule

\textbf{Metric} & $\vert \Delta d \vert $ & \multicolumn{3}{c}{\textbf{Pearson Coefficient}} \\
\cmidrule(lr){2-2} \cmidrule(lr){3-5}
 &\textbf{Weather} & \textbf{POIs} & \textbf{Road} & \textbf{Demographic} \\
\midrule

\textit{Top-1} & Snow (12.2$^\circ$) & Theatre (0.6390) & Trunk (0.5060) & Population With A Bachelor's Degree (0.5007) \\

\textit{Top-2} & Rain (8.62$^\circ$) & Clothes (0.5840) & Primary (0.4165) & Population With A Master's Degree (0.4914) \\

\textit{Top-3} & Haze (7.55$^\circ$) & Fast Food (0.5813) & Tertiary (0.3306) & Population With A Doctorate (0.3532) \\

\textit{Top-4} & Mist (4.75$^\circ$) & Cafe (0.5739) & Secondary (0.2970) & Population Above Poverty (0.3389) \\
\bottomrule
\end{tabular}
\end{center}
\end{table*}

\subsection{Selected Contextual Features}\label{revisitingcontextFeature}
We here elaborate on seven kinds of contextual features, including three temporal and four spatial contextual features.

\begin{figure}[h]
	\centering
	\includegraphics[width=.9\linewidth]{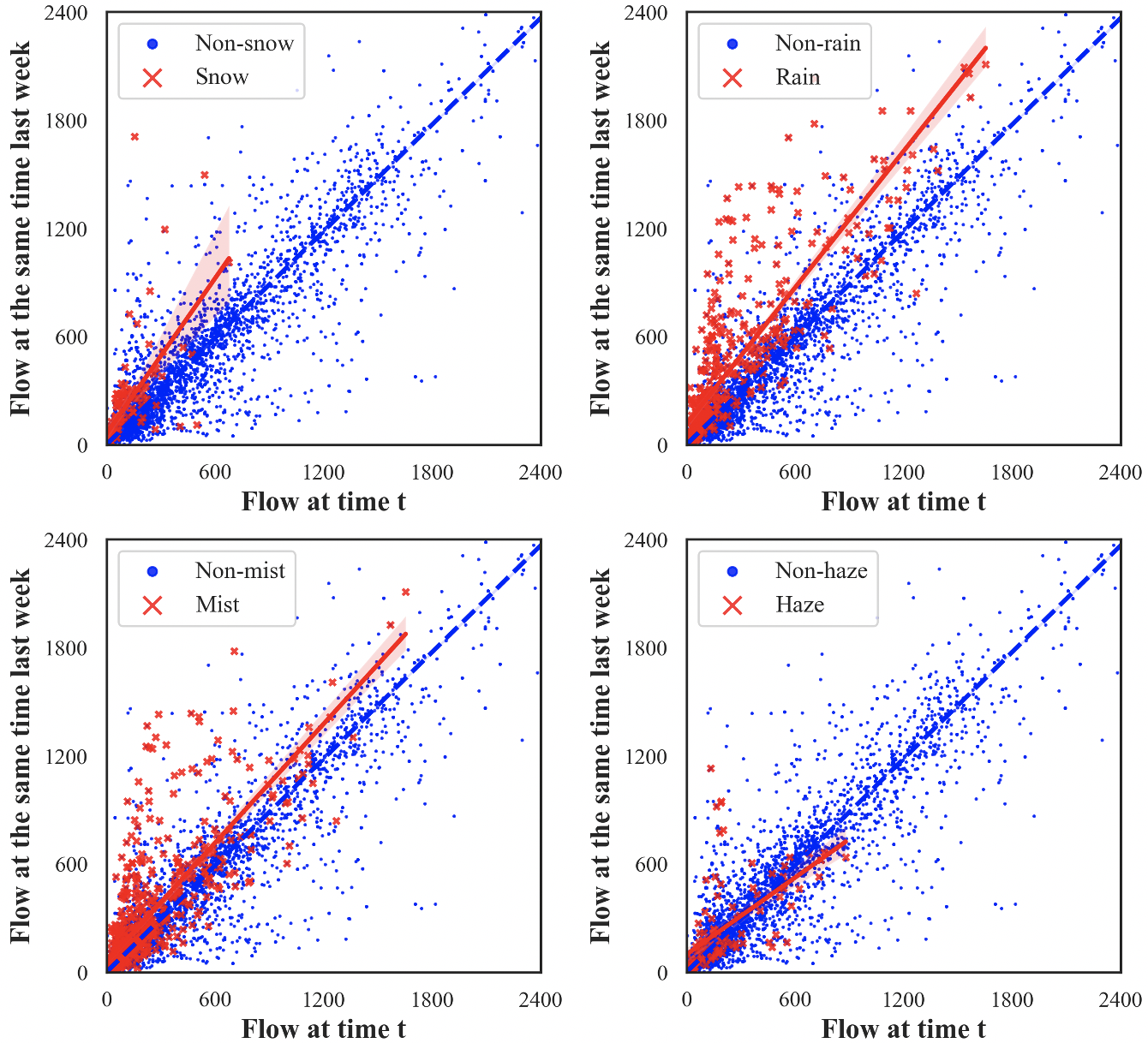}
    \vspace{-.5em}
	\caption{Impact of different weather conditions on citywide crowd flow. The red solid line represents a specific weather condition (e.g., snow), while the blue dashed line represents the other conditions in comparison (e.g., non-snow). The greater the difference between the red and blue curves, the more significant the impact of the weather conditions on crowd mobility. Best viewed in color.}
	\label{fig: weather_influence}
\end{figure}

\textbf{Weather.} In general, weather features such as temperature, humidity, wind speed, and weather states (e.g., cloudy and thunderstorms) are associated with short-term fluctuations in the atmosphere, which can impact crowd flow dynamics \cite{li_traffic_2015,hoang_fccf:_2016,wuInterpreting2016}. For example, a rise in temperature may result in increased use of bike-sharing services \cite{li_traffic_2015}, while heavy rains and strong winds may decrease the utilization of bike-sharing and online ride-hailing services \cite{hoang_fccf:_2016}. The impact of weather on crowd flow may vary depending on the specific application scenario of STCFP. Notably, temperature and weather states are the most commonly examined contextual features in existing literature, as indicated in the second column of Table \ref{tab:features}.

Our analysis confirms that snow and rain have a more significant impact on reducing crowd mobility compared to other weather conditions. Given the strong weekly periodicity in urban crowd mobility \cite{zhang2017deep, STMeta}, we compare crowd flows during specific weather conditions to those from previous weeks. To provide a macro-level evaluation of weather's impact, we aggregate flows across all stations to calculate the total citywide volume, reducing the crowd flow time series $\textbf{X} \in \mathbb{R}^{T \times N \times D}$ to $\mathbb{R}^{T \times D}$. 
In Figure~\ref{fig: weather_influence}, we plot the citywide crowd flow at time $t$ on the X-axis and the corresponding flow from the same time in the previous week on the Y-axis.
The red solid line represents a specific weather condition (e.g., snow at time $t$ but not at the same time last week), while the blue dashed line depicts other conditions (e.g., no snow at either time).
If a weather condition affects crowd mobility, the flow curve either steepens or flattens. For example, if rain reduces crowd mobility, the slope of the rain curve will be steeper than the curve under other weather conditions. We define $\vert \Delta d(S_n) \vert$ as the influence of weather condition $S_n$, calculated as the included angle between two curves:
\begin{equation} \vert \Delta d(S_n) \vert = \arctan(slope({S_n})) - \arctan(slope({\overline{S_n}})) 
\end{equation}
Here, $slope(S_n)$ represents the curve for weather condition $S_n$ (red lines in Figure~\ref{fig: weather_influence}), while $\overline{S_n}$ represents the curve for other weather conditions (blue lines). Table \ref{tab: correlation_score} lists the top four weather conditions with the largest $\vert\Delta d\vert$. It indicates that snow (12.2$^\circ$) and rain (8.62$^\circ$) have a more significant impact on reducing crowd mobility compared to other weather conditions.

\begin{figure}[h]
  \centering
  \begin{minipage}[c]{0.8\linewidth}
  \includegraphics[width=1\linewidth]{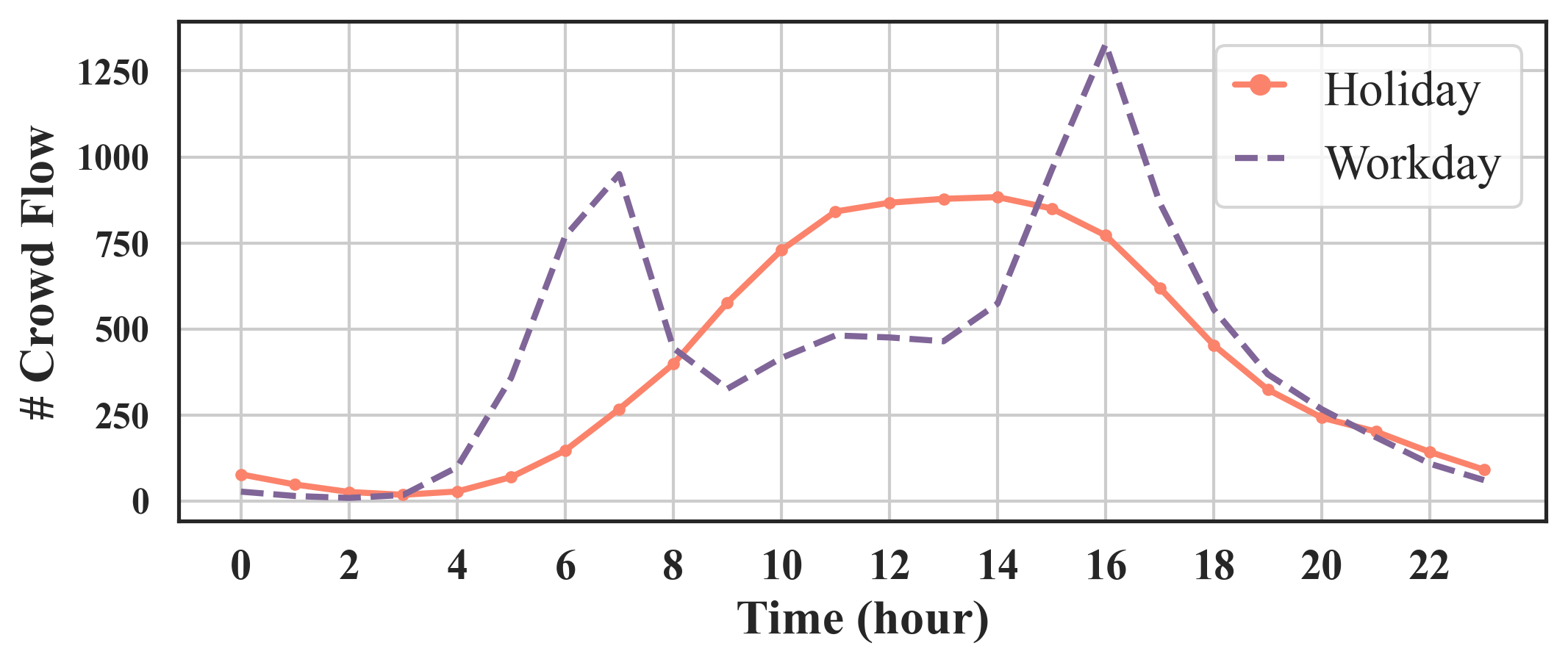}
  \vspace{-2.5em}
  \caption{Citywide hourly crowd flow during holidays and workdays.}
  \label{fig: holiday_workday_pattern}
  \end{minipage}
  \hspace{1mm}
  \begin{minipage}[c]{0.8\linewidth}
  \vspace{.5em}
  \includegraphics[width=1\linewidth]{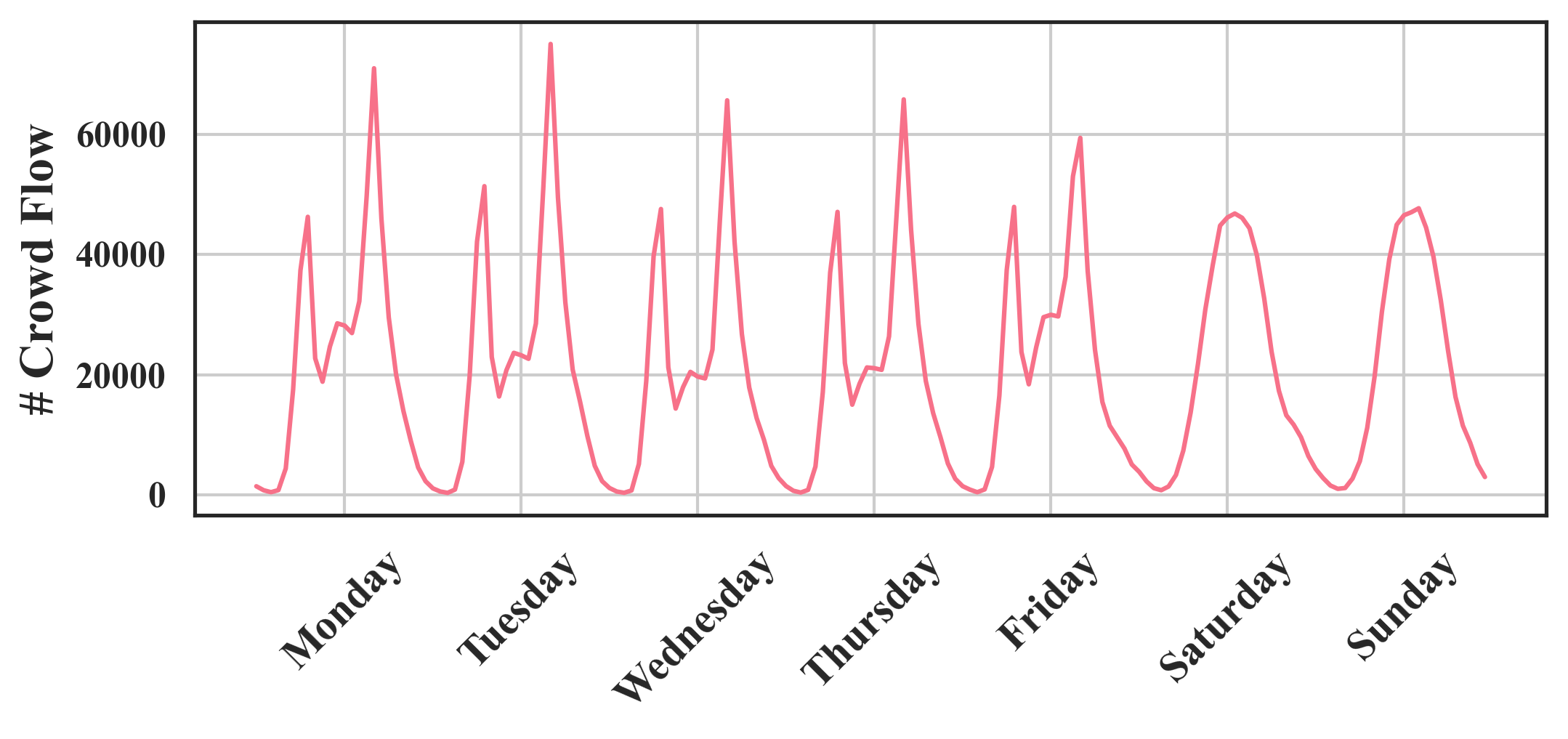}
  \vspace{-2.5em}
  \caption{Citywide daily crowd flow during a week.}
  \label{fig: week_pattern}
  \end{minipage}
\end{figure}

\textbf{Holiday.} During the holiday, people usually take a break from work or other regular activities in order to relax or travel. As a result, the crowd flow daily patterns are closely related to holidays. There is an obvious migration flow from urban residential areas to business areas during workdays, but this pattern is unclear on holidays \cite{zhang2017deep}. 
In Figure~\ref{fig: holiday_workday_pattern}, our analysis confirms that the citywide mobility pattern on weekends (one peak) is different from that on weekdays (two peaks), which is consistent with previous research findings \cite{hoang_fccf:_2016}.

\textbf{Temporal Position.} \label{TemporalPositionFeature} 
The temporal dynamics of crowd flow exhibit significant variation across different periods. Figure~\ref{fig: week_pattern} illustrates higher flow volumes on Friday evenings compared to Thursday evenings. Besides, hourly flow patterns during workdays can also differ, with peak flow volumes during morning and evening rush hours far exceeding those during off-peak periods, as shown in Figure~\ref{fig: holiday_workday_pattern}.
The above phenomena reveal the temporal heterogeneity of crowd flow \cite{STDM_problem_2018}. To capture these temporal variations in crowd flow, two discrete indicators are introduced to encode temporal position information \cite{liu_deeppf:_2019, yi2019citytraffic}:
\begin{itemize}
	\item \emph{HourofDay} indicates where the current time is in one day (what time it is), and its value ranges from 0 to 23 (representing 0 o'clock to 23 o'clock). 
	\item \emph{DayofWeek} indicates where the current time is in one week (which day it is). The value range of \emph{DayofWeek} is between Monday and Sunday. 
\end{itemize}
Temporal position features readily adjust to fine-grained STCFP tasks (e.g., 15-minute prediction) by encoding each time interval as a unique value within a day.

\textbf{Points Of Interests.} 
POIs are specific locations or landmarks that hold significance for people, such as residential areas, business areas, and tourist attractions. These can include attractions, historical sites, cultural centers, entertainment venues, or any other place that captures people's attention. POIs enhance our understanding of crowd flow patterns in various locations \cite{yuan2014discovering}, offering valuable insights for modeling spatial correlations \cite{tong_simpler_2017,liang_geoman:_2018, sthan_2022}.

Our analysis in Figure~14 (available in the online appendix) confirms a clear correlation between POIs and crowd flow data. Each plot in Figure~14 illustrates the correlation between crowd flow data and different types of POIs (e.g., shops and clothing). Points on the plots represent counts of specific POI types alongside corresponding crowd flow data at each location. Theater, clothing, and fast food exhibit strong spatial correlations, as indicated by their coefficients in Table \ref{tab: correlation_score}, which is consistent with previous research \cite{lin2019deepstn+}. Notably, POIs exhibit the highest Pearson coefficient among roads and demographics. As later shown in our experiments, only POIs show significant improvement among spatial contextual features. This suggests filtering out spatial contexts with low correlation may lead to robust and enhanced performance.

\textbf{Road Network.}
The road network is crucial for understanding crowd flow dynamics, revealing spatial and temporal patterns of vehicular movement. Its topology, encompassing connectivity between road segments, junctions, and intersections, governs urban traffic flow. Previous research reveals that the spatial correlation is dominated by road network structure, with connected roads sharing similar crowd flow patterns \cite{li2017diffusion, geng2019spatiotemporal}.
 
Besides, the road network structure, such as the number of high-level road segments, effectively complements traffic modeling \cite{ambulance_prediction_2021, fine_grained_2021}. Our analysis in Figure~15 confirms a potential correlation (Pearson coefficient is 0.5060) between crowd flow and high-level trunk roads.

\textbf{Demographic.}
Demographic features, including age, gender, income, and education, are crucial for understanding and predicting crowd mobility, influencing commuting patterns to various centers \cite{hu_evolving_2018}. Previous studies indicate that communities with lower average household incomes, higher proportions of essential workers, and higher minority populations exhibit increased mobility levels \cite{chen2022_covid_demographic}. Moreover, a higher ratio of older adults negatively correlates with per capita mobility, while ratios of essential workers and minorities positively correlate with per capita mobility \cite{chen2022_covid_demographic}. Our analysis in Figure~16 confirms a moderately positive correlation (Pearson coefficient is 0.5007) between crowd mobility and the population education level.

\textbf{Spatial Position.}
The crowd mobility distribution exhibits significant variation across different areas, known as spatial heterogeneity \cite{wang_STDM_survey}. To model such characteristics, spatial position, typically represented by geographic coordinates like latitude and longitude, uniquely identifies the predicted area. Similar to encoding temporal information, spatial coordinates (e.g., \textit{AreaID} or \textit{CoordodCity} \cite{wang_deepsd:_2017}) can encode spatial positioning information.

\subsection{Contextual Features Preprocessing Methods}
To enhance the exploitation of contextual features, we elaborate on typical feature preprocessing techniques for the selected contextual features. Our study also includes a detailed discussion on context modeling techniques, as introduced in Section~\ref{fusionMethod}. It is worth noting that these modeling techniques usually employ learning paradigms that involve learnable parameters and thus differ from the contextual preprocessing methods.

\textbf{Weather.} The weather context encompasses both numerical and categorical features, which necessitates consideration in machine learning approaches such as neural networks. Numerical features, such as temperature ($^{\circ}$F), humidity (\%), and wind speed (m/s), hold specific numerical values and can be readily utilized. In contrast, categorical contextual features, such as weather states (e.g., sunny, rainy, etc.), require transformation into numerical features. One-hot encoding and hand-crafted encoding that categorizes weather states into good (e.g., sunny, cloudy) and bad weather (e.g., rainy, stormy, dusty) can both be used for this purpose \cite{hoang_fccf:_2016,wang_dynamic_nodate}.
\begin{figure*}[htbp]
  \centering
  \begin{minipage}[c]{0.7\linewidth}
  \includegraphics[width=1\linewidth]{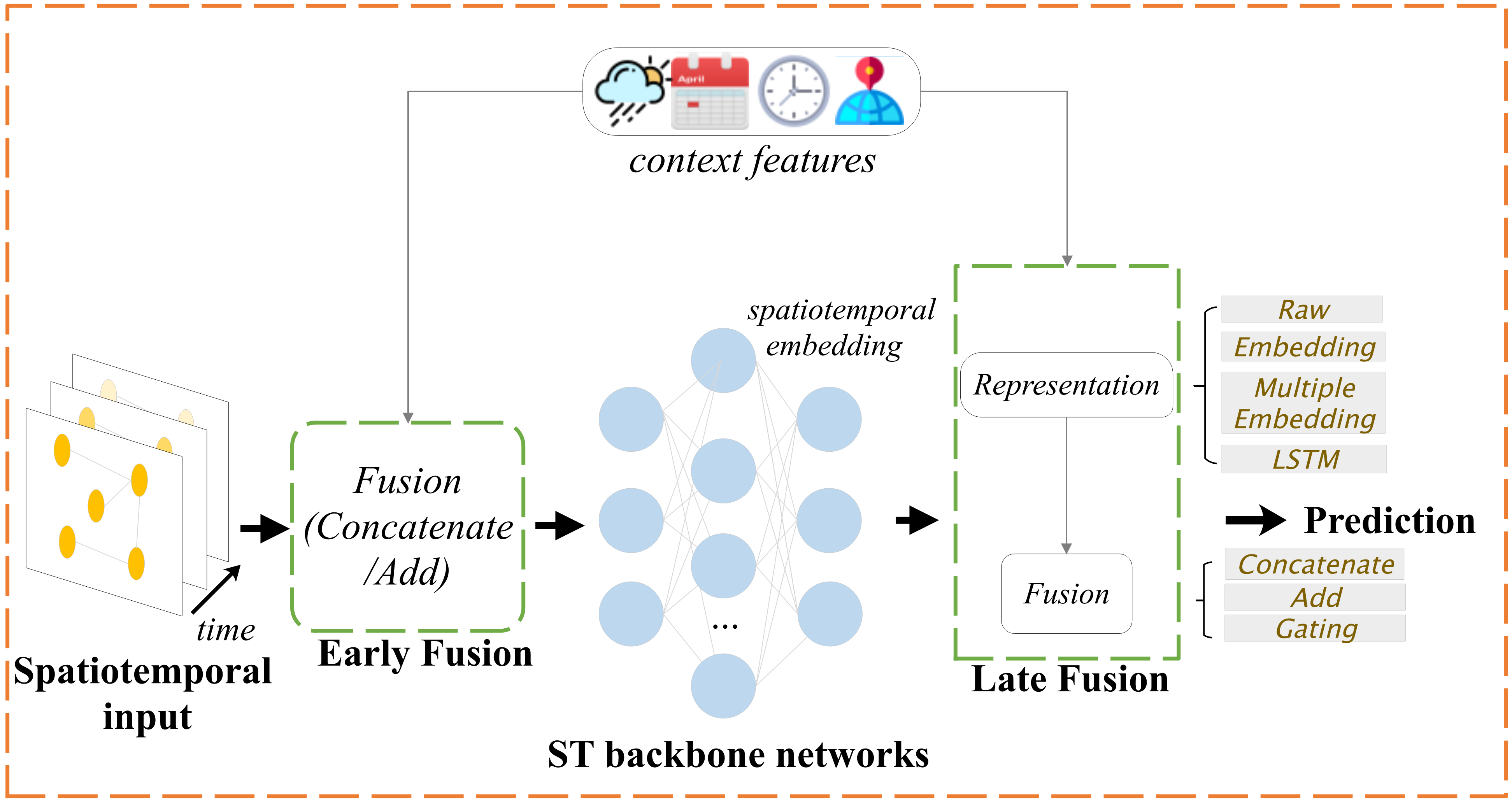}
  \caption{The proposed context modeling analytic framework. Existing modeling techniques can be categorized as Early Joint Modeling and Late Fusion. The backbone networks can be \textit{STMGCN}, \textit{STMeta}, or other spatiotemporal networks.}
  \label{framework}
  \end{minipage}
  \hspace{1mm}
  \begin{minipage}[c]{0.27\linewidth}
  \includegraphics[width=0.8\linewidth]{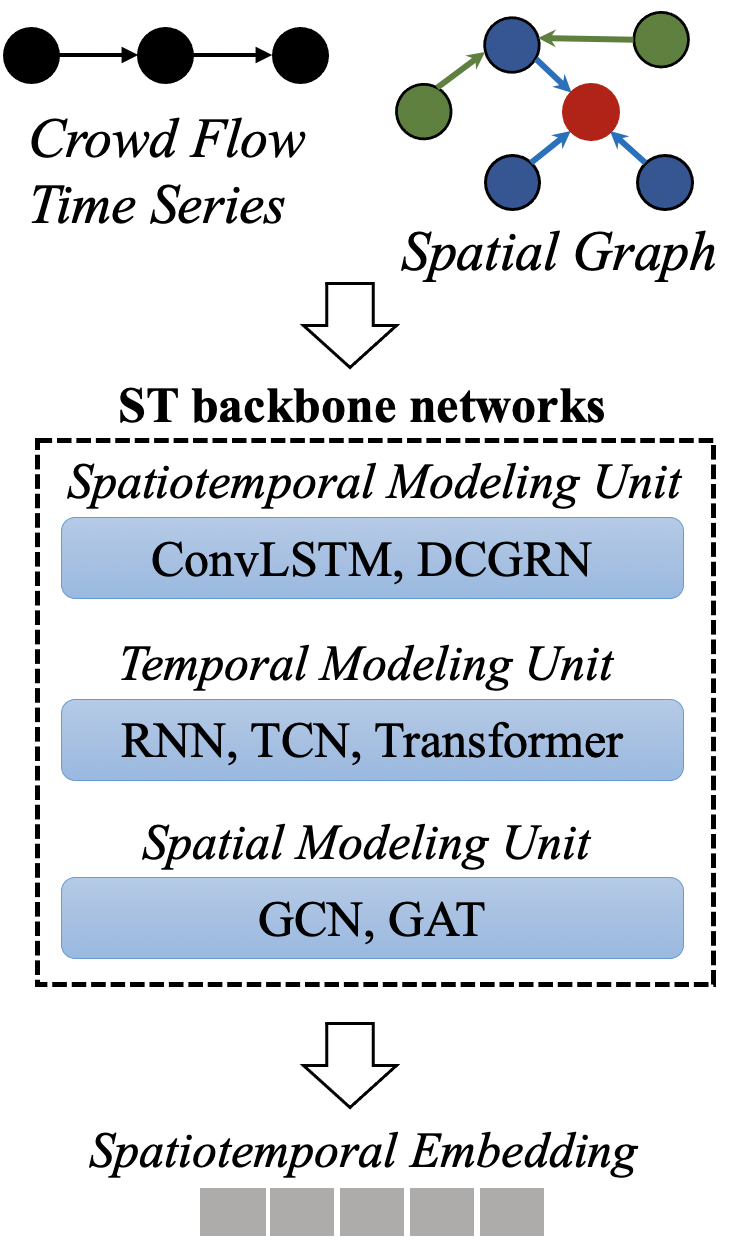}
  \caption{ST backbone networks consist of spatiotemporal modeling units or stacking of several temporal modeling units and spatial modeling units.}
  \label{st_backbone}
  \end{minipage}
\end{figure*}

\textbf{Holiday, Temporal Position and Spatial Position.} For the holiday context, we can use a binary variable to indicate whether a day is a holiday. Moreover, we also may include other features such as the type of holiday (e.g., Christmas, Thanksgiving). In a similar manner, the temporal and spatial position features can also be transformed into categorical one-hot vectors \cite{wang_deepsd:_2017, yi2019citytraffic, liu_deeppf:_2019}. For instance, \emph{HourofDay} can be represented as a one-hot vector of length 24, indicating the hour's position within a day. \emph{DayofWeek} can be encoded as a seven-dimensional vector. In a bike-sharing application with 288 stations, \textit{CoordofCity} is a 288-dimensional vector.

\textbf{POIs.} For POIs, researchers often count the POIs density of different categories as features \cite{tong_simpler_2017}. Due to the disturbances caused by various region functions, previous research proposes an elaborated feature engineering method to encode POIs features (i.e., the Inherent Influence Factor) \cite{zheng_deepstd_2020}. To differentiate the importance of each type of POIs to each region, previous research applies the Term Frequency–Inverse Document Frequency (\textbf{TF-IDF}) to encode POIs features, where the POIs are considered as words and the regions are the documents \cite{yuan_activity_2015, sthan_2022}.

\textbf{Road.} Road network data can be incorporated into crowd mobility prediction models through two main approaches. The first method produces a spatial graph based on the connectivity of the road network. This graph facilitates the integration of neighborhood information via adjacency relationships \cite{li2017diffusion, geng2019spatiotemporal}. The second approach processes the road network as geographical features, computing attributes such as road network types and densities within a given area \cite{travel_time_cooperative_2022}.

\textbf{Demographic.} 
Demographic data are collected within defined boundaries, such as census tracts or administrative districts. These boundaries serve as geographic units for aggregation and analysis. A common approach is to link demographic features to specific spatial areas based on corresponding administrative boundaries \cite{wang_subways_two_way_2022}. For example, census tract demographic data can be used as spatial features for bike stations within the tract.

Note that, in the above discussions, one-hot encoding has been widely used for transforming categorical features into numerical features, it can result in an explosion of feature dimensions, thereby increasing the risk of the curse of dimensionality. To address this issue, some researchers have manually reduced feature dimensions by categorizing weather states into good and bad weather categories. Others use \emph{Embedding} \cite{chai_multi_graph_2018,zhang2017deep} to achieve dimension reduction, which will be introduced in the next section.

\section{Analytical Studies on Context Modeling Techniques} \label{fusionMethod}
\subsection{Analytic Framework}
To illustrate the context modeling techniques more clearly, we first introduce a flexible and general context modeling analytic framework (Figure \ref{framework}). It contains two vital components, namely the ST backbone networks and the context modeling techniques. 

The ST backbone networks (Figure~\ref{st_backbone}) take spatiotemporal input (e.g., crowd flow time series and spatial graphs) and learn diverse spatiotemporal embeddings, having been well-studied in the previous research \cite{zhang2017deep,co_prediction_2019,meta_learning_traffic_2019,AGCRN_2020,yuan_demand_2021,ode_traffic_2021,STMeta}. Typically, ST backbone networks can be simplified to a function $g$ :
\begin{equation}
    g: \mathcal{X}_{(t-P):t} \in \mathbb{R}^{P\times N \times D} \rightarrow \mathcal{X}^{emb}_{t} \in \mathbb{R}^{N \times D_1} \label{eq:st_backbone}
\end{equation}
where $D_1$ is the number of features of spatiotemporal embedding. ST backbone networks extract spatiotemporal embedding by stacking several temporal modeling units (e.g., \textit{LSTM} \cite{hochreiter1997long}, \textit{TCN} \cite{gated_tcn_2017}, and \textit{Transformer} \cite{xu2020spatial}) and spatial modeling units (e.g., \textit{GCN} \cite{kipf2017semisupervised, chai_multi_graph_2018} and \textit{GAT} \cite{velivckovic2017graph,GMAN_AAAI2020}). Previous research also has proposed several spatiotemporal modeling units that can simultaneously model spatiotemporal dependencies (e.g., \textit{ConvLSTM}~\cite{convlstm_2015} and \textit{DCGRU}~\cite{li2017diffusion}).

Our analytic framework classifies existing context modeling techniques into \textbf{Early Joint Modeling} and \textbf{Late Fusion}. Early joint modeling fuses contextual features with spatiotemporal input before capturing spatiotemporal dependencies, while late fusion fuses contextual features with spatiotemporal embedding learned by ST backbone networks. The main difference is that \textbf{Early Joint Modeling} uses the same spatiotemporal backbone networks to extract both spatiotemporal embedding and context embedding while \textbf{Late Fusion} uses distinct modules to learn context embedding.

Note that our analytic framework is developed mainly for deep learning models (we summarize the context modeling techniques in Table \ref{tab:features}). Traditional statistical learning models such as \textit{XGBoost} \cite{chen2016xgboost} usually directly fuse contextual features by concatenating \cite{li_traffic_2015,yang_mobility_2016,tong_simpler_2017,LiBikeTKDE2019}.

\subsection{Early Joint Modeling} \label{early_joint_modeling}
Early joint modeling refers to fusing contextual features with raw spatiotemporal inputs before capturing spatiotemporal dependencies via ST backbone networks. Hence, in addition to getting spatiotemporal embedding of spatiotemporal inputs, the ST backbone networks will also extract the spatiotemporal dependencies of contextual features synchronously.
For example, researchers combine the features of \emph{DayofWeek}, \emph{HourofDay} and \textit{POIs' population distribution map} by adding and then applying the ResPlus structure to capture spatial dependencies between distant locations \cite{lin2019deepstn+}. Yao et al. first concatenate spatial features and contextual features (including weather and holidays) and then capture temporal patterns by \textit{LSTM} \cite{yao2018DeepMulti,Yao2018ModelingSD}. 

\begin{figure*}[htbp]
  \centering
  \begin{minipage}[c]{0.58\linewidth}
  \includegraphics[width=1\linewidth]{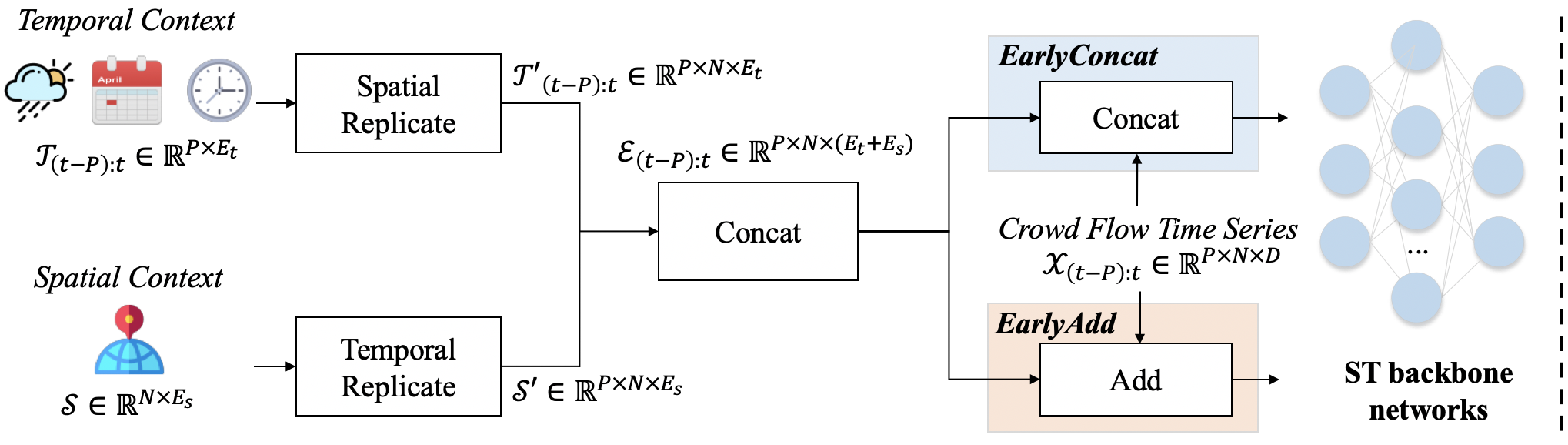}
  \caption{Details of early joint modeling.}
  \label{fig:earlyfusion}
  \end{minipage}
  \hspace{1mm}
  \begin{minipage}[c]{0.395\linewidth}
  \includegraphics[width=1\linewidth]{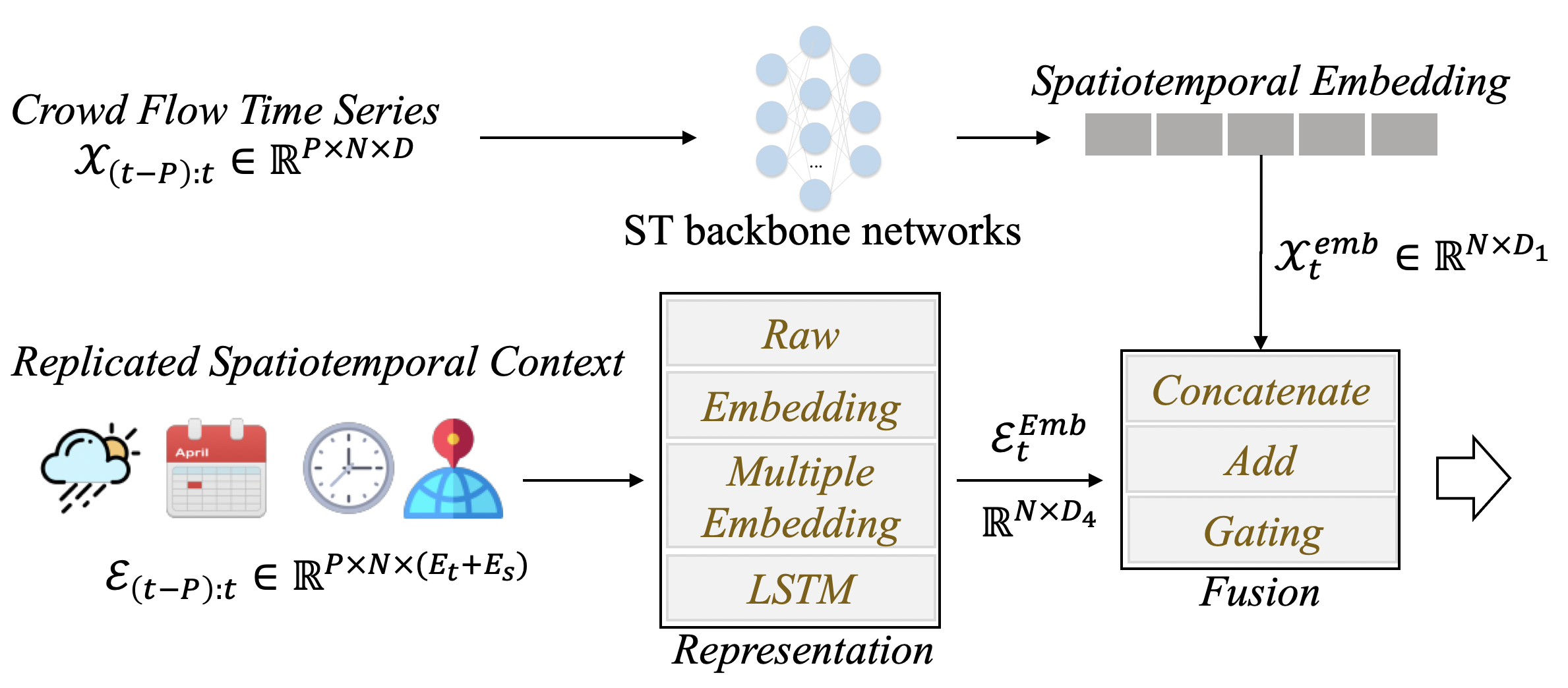}
  \caption{Details of late fusion.}
  \label{fig:latefusion}
  \end{minipage}
\end{figure*}

Notably, weather, holiday, and temporal position are temporal contextual features (i.e., $\mathcal{T}_{(t-P):t}\in \mathbb{R}^{P\times E_t}$) and POIs are spatial contextual features (i.e., $\mathcal{S} \in \mathbb{R}^{N\times E_s}$). To align their feature dimension with crowd flow time series (i.e., $\mathcal{X}_{(t-P):t}\in \mathbb{R}^{P\times N \times D}$), \textit{replicate technique} (details in Figure~\ref{fig:earlyfusion}) is widely used \cite{ke_short-term_2017,zhang_flow_2019,regiontrans_2019}. After replicating, the duplicated temporal and spatial contextual features are $\mathcal{T}^{\prime}_{(t-P):t}\in \mathbb{R}^{P\times N \times E_t}$ and $\mathcal{S}^{\prime} \in \mathbb{R}^{P \times N\times E_s}$, respectively. As the semantic meaning of features dimension (i.e., the last axis) is the same, for brevity, we merge them into the spatiotemporal contextual features $\mathcal{E}_{(t-p):t}\in \mathbb{R}^{P\times N\times (E_t+E_s)}$ by concatenating.

As shown in Figure~\ref{fig:earlyfusion}, both \emph{Concatenate} and \emph{Add} can fuse spatiotemporal contextual features $\mathcal{E}_{(t-P):t}$ and crowd flow time series features $\mathcal{X}_{(t-P):t}$, and we name these two modeling techniques \emph{EarlyAdd} and \emph{EarlyConcat}, respectively. Recall that $P$ is the number of historical observations, $N$ is the number of locations, $E_s+E_t$ and $D$ are the dimensions of contextual features and spatiotemporal inputs. We denote the output of early joint modeling as $\mathcal{O}^{Early}$ and the formula of \emph{EarlyConcat} is:
\begin{equation}
\mathcal{O}^{Early}_{(t-P):t} = Concat(\mathcal{E}_{(t-P):t};\mathcal{X}_{(t-P):t}) \in \mathbb{R}^{P\times N \times (D+E_s+E_t)}
\end{equation}
Similarly, the output of early joint modeling given by \emph{EarlyAdd} is:
\begin{equation}
\mathcal{O}^{Early}_{(t-P):t} = \mathcal{E}_{(t-P):t} \times W_e + \mathcal{X}_{(t-P):t} \times W_{st} + b_{e} \in \mathbb{R}^{P\times N\times D2}
\end{equation}
where $W_e \in \mathbb{R}^{(E_s+E_t) \times D_2}$, $W_{st} \in \mathbb{R}^{D \times D_2}$ and $b_{e} \in \mathbb{R}^{D_2}$ are trainable parameters. $D, D_2$ are the dimensions of the crowd flow time series and the fused features obtained by \emph{EarlyAdd}, respectively. 

Then the ST backbone networks can jointly model spatiotemporal dependencies of both crowd flow series and contextual features. Equation~\ref{eq:st_backbone} can be rewrite as:
\begin{equation}
g^{\prime}: \mathcal{O}^{Early}_{(t-P):t} \in \mathbb{R}^{P\times N \times D^{\prime}} \rightarrow \mathcal{X}^{emb}_{t} \in \mathbb{R}^{N \times D_3} 
\end{equation}
where $D_3$ is the dimension of joint embedding of crowd flow series and contextual features. The joint embedding usually is fed to the output layer and gives predictions. $D^{\prime}$ is $D$+$E_s$+$E_t$ or $D_2$ when applied \emph{EarlyConcat} and \emph{EarlyAdd}, respectively.

\subsection{Late Fusion} \label{late_fusion_modeling}
Late fusion combines contextual features and spatiotemporal embedding learned by ST backbone networks, which is adept at fusing data from different domains \cite{zhang2016dnn}. In this study, we investigate existing late fusion methods and identify two steps in the process. The first step involves the learning of diverse contextual embedding, while the second step entails fusing the context embedding with spatiotemporal embedding for predictions. This two-step approach is a common thread among the late fusion methods. These findings contribute to the understanding of the late fusion technique and provide insights for its future development and applications in the STCFP problem. We next elaborate on these two steps.

\subsubsection{Context Representation Step}
The representation step takes in $\mathcal{E}_{(t-p):t}$, the replicated spatiotemporal contextual features\footnote{The replicate techniques details are introduced in Figure~\ref{fig:earlyfusion}.}, and outputs the learned context embedding $\mathcal{E}^{Emb}_t\in\mathbb{R}^{N\times D_4}$. The following four techniques can be applied to conduct transformation as shown in Figure~\ref{fig:latefusion}.

\textbf{\emph{Raw}}. No transformations are applied to the spatiotemporal contextual features. Since the contextual features that are closer in time to the prediction moment should be more important, \emph{Raw} keeps the contextual features closest to the prediction moment, namely $\mathcal{E}_{t-1}$.

\textbf{\emph{Embedding}}. The embedding technique is widely used in various fields, such as NLP (Natural Language Processing). It maps sparse high-dimension features to dense low-dimension features. Many studies use fully connected layers for embedding \cite{chai_multi_graph_2018,zhang2017deep}. For instance, Zhang et al. \cite{zhang2017deep} use two fully connected layers upon contextual features, and the first layer acts as an embedding layer. \emph{Embedding} keeps the closest contextual features $\mathcal{E}_{t-1}$ and the embedding technique is applied to the feature dimension of $\mathcal{E}_{t-1}$ (i.e., the last axis).

\textbf{\emph{Multiple Embedding}}. Different contextual features can be fed into multiple embedding layers \cite{liu_deeppf:_2019,liang_geoman:_2018,wang_deepsd:_2017}. For example, Liu et al. \cite{liu_deeppf:_2019} use different embedding layers to model weather and temporal position, respectively. \emph{Multiple Embedding} also uses the closest contextual features $\mathcal{E}_{t-1}$ and transforms the feature dimension of $\mathcal{E}_{t-1}$ (i.e., the last axis).

\textbf{\emph{LSTM}}. Temporal contextual features (e.g., weather) usually are time-varying, and past contextual features may also impact future flow. For example, sudden heavy rain may immediately reduce the crowd flow, but when the rain stops, the crowd flow may be larger than ever. Ke et al. \cite{ke_short-term_2017} use LSTM to capture temporal dependencies of weather. Unlike the above three representation techniques, which focus on the closest contextual features $\mathcal{E}_{t-1}$, \emph{LSTM} takes in the historical context series $\mathcal{E}_{(t-P):t}$.
	
\subsubsection{Feature Fusion Step} 
\begin{figure}[h]
\centering
 \includegraphics[width=.99\linewidth]{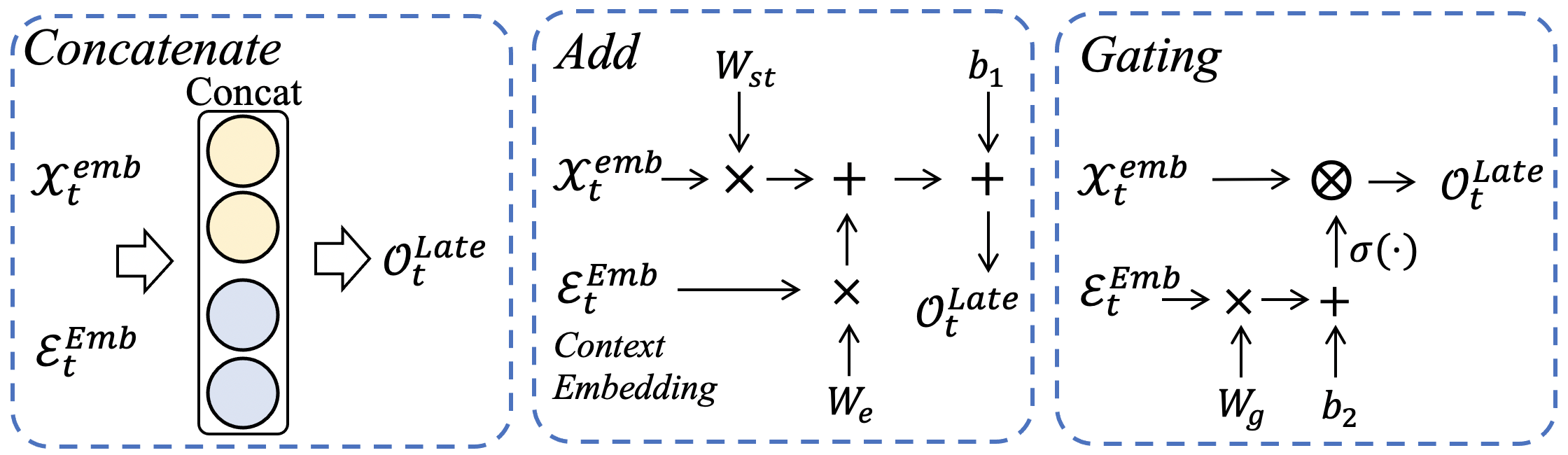}
\caption{Details of \emph{Concatenate}, \emph{Add} and \emph{Gating}.}
\label{fusion_details}
\end{figure}

The fusion step fuses the context embedding $\mathcal{E}^{emb}_t\in \mathbb{R}^{N\times D_4}$ with the spatiotemporal embedding $\mathcal{X}^{emb}_t\in \mathbb{R}^{N\times D_1}$ and outputs the fused embedding $\mathcal{O}_t^{Late}$ for prediction. The following three techniques are widely adopted in previous research (details are in Figure~\ref{fusion_details}).

\textbf{\emph{Concatenate}} is a widely used technique that combines different features. Chai et al. \cite{chai_multi_graph_2018} use a fully connected layer as the embedding layer to represent features and then concatenate $\mathcal{E}^{emb}_t$ with spatiotemporal embedding $\mathcal{X}^{emb}_t$. The \emph{Concatenate} formula is:
\begin{equation}
\mathcal{O}^{Late}_t = Concat(\mathcal{X}^{emb}_t;\mathcal{E}^{emb}_t) \in \mathbb{R}^{N\times (D_1+D_4)}
\end{equation}

\textbf{\emph{Add}} first projects the context embedding $\mathcal{E}^{emb}_t$ and the spatiotemporal embedding $\mathcal{X}^{emb}_t$ into the feature space with the same dimension. Then it fuses them by adding the corresponding aligned features. The formula of \emph{Add} is listed below. $W_{st}\in \mathbb{R}^{D_1 \times D_5}$, $W_e\in\mathbb{R}^{D_4 \times D_5}$, and $b_1 \in \mathbb{R}^{D_5}$ are learnable parameters.
\begin{equation}
\mathcal{O}^{Late}_t = \mathcal{X}^{emb}_t \times W_{st} + \mathcal{E}^{emb}_t \times W_e + b_1
\end{equation}

\begin{table}[t]
	\small
	\caption{\textit{Late Fusion} includes representation and fusion steps. Modeling techniques with `*' are newly found based on our analytic studies, which have never appeared in the literature to the best of our knowledge.}
    \vspace{-2em}
	\label{context_modeling}
	\begin{center}
  \resizebox{0.49\textwidth}{!}{
			\begin{tabular}{lccc}
				\toprule
\textbf{Name}&\textbf{Representation}&\textbf{Fusion}&\textbf{Literature}\\
\midrule
\emph{Raw-Concat}& \multirow{3}{*}{\textit{Raw}}&\textit{Concatenate}&\cite{zhu_deep_2017}\\
\emph{Raw-Add}&&\textit{Add}&\cite{zhang2016dnn,zonoozi2018}\\
\emph{Raw-Gating}&&\textit{Gating}&\cite{zhang_flow_2019}\\
\midrule
\emph{Emb-Concat}& \multirow{3}{*}{\textit{Embedding}}&\textit{Concatenate} & \cite{chai_multi_graph_2018,yuan_demand_2021}\\
\emph{Emb-Add}&&\textit{Add}&\cite{zhang2017deep,yi2019citytraffic}\\
\emph{Emb-Gating}&&\textit{Gating} & \cite{sunIrregular}\\
\midrule
\emph{MultiEmb-Concat}& \multirow{3}{*}{\shortstack{\textit{Multiple} \\ \textit{Embedding}}} &\textit{Concatenate}& \cite{liu_deeppf:_2019,liang_geoman:_2018,wang_deepsd:_2017,Zhang_Huang_Xu_Xia_Dai_Bo_Zhang_Zheng_2021}\\
\emph{MultiEmb-Add$^*$}&&\textit{Add}&-\\
\emph{MultiEmb-Gating$^*$}&&\textit{Gating}&-\\
\midrule
\emph{LSTM-Concat$^*$}& \multirow{3}{*}{\emph{LSTM}}&\textit{Concatenate}& - \\
\emph{LSTM-Add}&&\textit{Add}& \cite{ke_short-term_2017}\\
\emph{LSTM-Gating$^*$}&&\textit{Gating}&-\\
				\bottomrule
		\end{tabular}}
	\end{center}
\vspace{-1.5em}
\end{table}

\textbf{\emph{Context-gating}}~\cite{zhang_flow_2019} regards the contextual features as the activation function of crowd flow. It first transforms the contextual embedding to the gating value $G$ and then uses $G$ to activate the spatiotemporal embedding:
\begin{gather}
G=\sigma (\mathcal{E}^{emb}_t \times W_{g} + b_2) \in \mathbb{R}^{N\times D_1} \\
\mathcal{O}^{Late}_t = \sigma (G \otimes \mathcal{X}^{emb}_t)
\end{gather}
where ${W}_{g}\in\mathbb{R}^{D4\times D1}$ and $b_2\in\mathbb{R}^{D_1}$ are learnable parameters, `$\otimes$' is the Hadamard product. $\sigma(\cdot)$ is the sigmoid function. The intuition of \emph{context-gating} is that features are like switches, and the crowd flows would be tremendously changed if a certain switch is activated. As gating mechanisms show great capability in other fields like natural language processing\cite{seq2seq} and computer vision~\cite{SENet2018}, we here briefly discuss their differences.

\textbf{Temporal-gating and channel-gating mechanisms}: In LSTM (Long Short-Term Memory), temporal gating mechanisms control the flow of information through the network (i.e., by the input gate, forget gate, and output gate) by selectively updating and exposing information in the cell state through these gates, enabling LSTM to learn temporal dependencies. In SENet (Squeeze-and-Excitation Network)~\cite{SENet2018}, channel-gating mechanisms perform global average pooling to compute channel-wise feature recalibration to produce a scaling vector that dynamically adjusts the channel importance. 

Even though context-gating mechanisms share similarities with LSTM and SENet, such as recalibrating feature embeddings, they still have distinct differences. Unlike LSTM, which relies on past features as gating's input, and SENet, which takes in the feature map itself, context-gating mechanisms\footnote{Without incurring ambiguity, we use gating mechanisms to refer to context-gating mechanisms in the rest of this paper.} take extra heterogeneous data sources (i.e., contextual features) as inputs. This broader data intake enables them to incorporate diverse information.

In summary, by combining the above two steps, we list 12 (3*4) variants of Late Fusion Modelling techniques, as shown in Table \ref{context_modeling}. It is worth noting that, with a comprehensive survey of the literature, we find that four variants never appeared in previous work. Meanwhile, we believe that these variants are reasonable context modeling techniques and we will also test them in our benchmark experiments.

\subsection{Discussion on Modeling Techniques} \label{assumption_discussion}
\textbf{Assumption on Modeling Stages}: The difference between early joint modeling and late fusion lies in their assumption of spatiotemporal patterns in contextual features compared to crowd flow series. Early joint modeling assumes similarity between these patterns, allowing for the same network for modeling, while late fusion posits differences, necessitating separate networks. While some research has identified correlations between traffic and contextual features \cite{li2017diffusion, lin2019deepstn+, chen2022_covid_demographic}, current spatiotemporal backbones are typically tailored toward crowd flow series rather than context. For instance, spatial relationships (e.g., functionality graphs~\cite{geng2019spatiotemporal}) are modeled based on traffic similarities without considering potential variations in weather conditions. Therefore, if significant disparities exist between context and crowd flow patterns in the application, early joint modeling may not produce optimal results. Late fusion could be more effective under such circumstances by employing distinct network modules to capture contextual representations.

\textbf{Assumption on Representation Techniques}: In the representation step, \textit{LSTM} variants consider past contexts' influence on future traffic predictions, whereas other variants (i.e., \textit{Raw}, \textit{Emb}, and \textit{Multi-Emb}) focus solely on present-time contexts affecting predicted flows. Compared to directly handling raw features without transformation, \textit{Emb} treats high-dimensional contexts through dimensionality reduction before subsequent learning due to dimensionality concerns. \textit{Multi-Emb} suggests that diverse context types have semantic distinctions requiring separate embedding for dimensionality reduction.

\textbf{Assumption on Fusion Techniques}: In the fusion step, \textit{Concat} variants concatenate features by assuming low correlation levels among them since highly correlated features might introduce redundant information impacting model generalization capabilities \cite{feature_selection_2003}. \textit{Add} variants assume that crowd flow representations and contextual embeddings share similar semantics, making them additive in vector. \textit{Gating} variants consider that context influences changes in crowd flow (e.g., heavy rain affecting bike-sharing usage like a switch \cite{zhang_flow_2019}) hence employing gating mechanisms mapping context into scaling factors acting upon crowd flow embeddings.

\section{Empirical Benchmark Setup} \label{BenchmarkDescription}
\subsection{Datasets} 
We collect six crowd mobility datasets for five tasks, including bike flow, traffic speed, metro passenger flow, electric vehicle charging, and pedestrian count prediction. We chose these tasks and datasets for two main reasons. Firstly, these five tasks cover common modes of transportation in cities (including walking, biking, metro, and cars) and are widely popular in existing research and benchmark studies \cite{dl_traffic_2021, libCity_2021} (e.g., bike flow and traffic speed tasks). Regarding the datasets chosen, we choose them because of their accessibility and widespread use in related research (e.g., Bike NYC \cite{zhang2017deep} and PEMS BAY \cite{li2017diffusion}. 

Corresponding to the crowd mobility datasets, we collect contextual data such as weather, holidays, POIs, roads, and demographics. The original crowd mobility records are processed at intervals of 30, 60, and 120 minutes. Moreover, these datasets are processed and publicly accessible in our benchmark repository. For detailed dataset descriptions, please refer to Appendix~A (available in the online supplemental material).

\subsection{Model Variants} We implement \textit{STMGCN} \cite{geng2019spatiotemporal} and \textit{STMeta} \cite{STMeta} as the spatiotemporal backbone networks in Figure \ref{framework}, since these two STCFP models have been verified to perform generally well in a recent large-scale benchmark study \cite{STMeta}. \emph{Early Joint Modeling} and \emph{Late Fusion} can be applied to both backbone networks. In addition, we implement an XGBoost-based predictive model \cite{chen2016xgboost}, by concatenating spatiotemporal and contextual features to further analyze the generalizability of contextual features on traditional machine learning models.

\subsection{Implementation Details}
Due to page limitations, we describe our implementation details and the experiment platform in Appendix B (available in the online supplemental material). 

\subsection{Evaluation Metrics} \label{comparingTestMethod} We exploit two widely used metrics, namely RMSE (Root Mean Square Error) and MAE (Mean Absolute Error) to assess the performance of each method:
\begin{align}
\operatorname{RMSE}(y,\hat{y})=\sqrt{\frac{1}{N}\sum_{i=1}^{N}(y_i-\hat{y_i})^2} \\
\operatorname{MAE}(y,\hat{y})=\frac{1}{N}\sum_{i=1}^{N}\lvert(y_i-\hat{y_i})\lvert
\end{align}

where $y_i$ and $\hat{y_i}$ are the ground truth and predicted flows and $N$ is the number of samples. Suppose that there is a set of approaches $\mathcal{X}$ and several evaluation datasets $\mathcal{D}$, \emph{avgNRMSE} and \emph{avgNMAE} are defined to assess the overall performance of each method. The \emph{avgNRMSE} and \emph{avgNMAE} of method $x\ (x \in \mathcal{X})$ are:
\begin{align}
avgNRMSE_{x}=\text {Average}_{d \in \mathcal{D}}\left(\frac{\operatorname{RMSE}_{x, d}}{\min _{x^{\prime} \in \mathcal{X}}\left(\operatorname{RMSE}_{x^{\prime}, d}\right)}\right) \\
avgNMAE_{x}=\text {Average}_{d \in \mathcal{D}}\left(\frac{\operatorname{MAE}_{x, d}}{\min _{x^{\prime} \in \mathcal{X}}\left(\operatorname{MAE}_{x^{\prime}, d}\right)}\right)
\end{align}
If $avgNRMSE_{x}$ and $avgNMAE_{x}$ are closer to 1, method $x$ exhibits superior performance across diverse datasets, suggesting better generalizability. The MAE results are consistent with the RMSE results. We list the RMSE results in the main manuscripts, while the MAE results are in Appendix E (available in the online supplemental material).

\section{Benchmark Results and Analysis} \label{sec: experiment_result}

\subsection{Results and Analysis on Modeling Techniques} 
To study the generalizability of context modeling techniques, we compare 14 techniques elaborated in Section \ref{fusionMethod}. The results are listed in Table \ref{tab: techniquues_60_STMeta_STMGCN_RMSE}, where we calculate \emph{avgNRMSE} to evaluate generalizability. 

We run each method at least three times and conduct \textit{t-test} to determine significant improvements. Techniques with significantly better \textit{avgNRMSE} than \textit{No Context} are highlighted in blue. We observe that only a few techniques have greater \emph{avgNRMSE} than \textit{No Context}, which demonstrates most techniques are with less generalizability, which further emphasizes the importance of investigating the generalizability of modeling techniques. 
Notably, only \emph{Raw-Gating} consistently outperforms \textit{No Context} with significantly better \emph{avgNRMSE} across two backbone networks (i.e., \textit{STMeta} and \textit{STMGCN}), showing its great generalizability.

\begin{table*}[htbp]
	\footnotesize
	\caption{60-minute RMSE results of different modeling techniques based on \textit{STMeta} and \textit{STMGCN}. The best results are highlighted in bold. \emph{No Context} does not incorporate any contextual features. The modeling techniques marked with * significantly ($p < 0.05$) outperform \emph{No Context}. Those with significantly better \textit{avgNRMSE} than \emph{No Context} are highlighted in {\color{blue!70}blue}.}
    \vspace{-2em}
	\label{tab: techniquues_60_STMeta_STMGCN_RMSE}
	\renewcommand\tabcolsep{4.0pt} 
	\begin{center}
    \begin{tabular}{lccccccccccccccccccccccccccccccccc}
	\toprule
	& \multicolumn{4}{c}{\textbf{STMeta}} & \multicolumn{4}{c}{\textbf{STMGCN}}\\
	\cmidrule(lr){2-5}  \cmidrule(lr){6-9} 
			
	\multicolumn{1}{c}{} & \multicolumn{1}{c}{\textit{Bike}} & \multicolumn{1}{c}{\textit{Metro}} & \multicolumn{1}{c}{\textit{EV}} & \multicolumn{1}{c}{\textit{avgNRMSE}} & \multicolumn{1}{c}{\textit{Bike}} & \multicolumn{1}{c}{\textit{Metro}} & \multicolumn{1}{c}{\textit{EV}} & \multicolumn{1}{c}{\textit{avgNRMSE}} \\
	\midrule
\textit{No Context} & 2.699$\pm$0.036 & 155.9$\pm$11.6 & 0.814$\pm$0.006 & 1.051$\pm$0.027  & 2.864$\pm$0.034 & 188.5$\pm$4.21 & 0.833$\pm$0.011 & 1.056$\pm$0.023 \\
\midrule
\multicolumn{2}{l}{\textbf{Early Joint Modeling}} \\
\textit{EarlyConcat} & 3.077$\pm$0.045 & 244.0$\pm$3.10 & 1.428$\pm$0.108 & 1.562$\pm$0.333 & 2.955$\pm$0.034 & 194.4$\pm$6.37 & 0.844$\pm$0.016 & 1.082$\pm$0.040 \\
\textit{EarlyAdd} & 2.804$\pm$0.095 & 210.2$\pm$24.1 & 0.814$\pm$0.026 & 1.187$\pm$0.205 & 2.778$\pm$0.066 & 188.4$\pm$2.80 & 0.850$\pm$0.021 & 1.052$\pm$0.018 \\
\midrule
\multicolumn{2}{l}{\textbf{Late Fusion}} \\
\textit{Raw-Concat} & 2.673$\pm$0.009 & 184.6$\pm$9.97 & 0.798$\pm$0.018 & 1.105$\pm$0.125  & 2.826$\pm$0.053 & 180.9*$\pm$2.00 & 0.874$\pm$0.028 & 1.053$\pm$0.036 \\
\textit{Raw-Add} & 2.636*$\pm$0.016 & 164.2$\pm$6.76 & 0.856$\pm$0.071 & 1.079$\pm$0.061  & 3.099$\pm$0.044 & 187.8$\pm$1.07 & 0.864$\pm$0.027 & 1.096$\pm$0.054 \\
\textit{Raw-Gating} & 2.612*$\pm$0.014 & 153.4$\pm$7.31 & 0.785*$\pm$0.004 & \cellcolor{blue!10}\textbf{1.022*$\pm$0.017} & 2.709*$\pm$0.017 & 193.2$\pm$2.80 & 0.816*$\pm$0.006 & \cellcolor{blue!10}1.029*$\pm$0.048 \\
\textit{Emb-Concat} & 2.666$\pm$0.031 & 172.9$\pm$2.11 & 0.784*$\pm$0.002 & 1.072$\pm$0.086  & 2.941$\pm$0.017 & 183.8$\pm$4.10 & 0.844$\pm$0.008 & 1.061$\pm$0.036 \\
\textit{Emb-Add} & 2.665$\pm$0.031 & 172.3$\pm$10.5 & 0.783*$\pm$0.010 & 1.070$\pm$0.084  & 2.818$\pm$0.042 & 189.8$\pm$3.06 & 0.840$\pm$0.003 & 1.055$\pm$0.023 \\
\textit{Emb-Gating} & 2.616*$\pm$0.023 & 157.4$\pm$8.88 & 0.788*$\pm$0.004 & 1.033$\pm$0.045  & \textbf{2.689*$\pm$0.014} & \textbf{181.7$\pm$7.33} & \textbf{0.808*$\pm$0.005} & \cellcolor{blue!10}\textbf{1.011*$\pm$0.018} \\
\textit{MultiEmb-Concat} & 2.648$\pm$0.062 & 173.8$\pm$7.44 & 0.789*$\pm$0.004 & 1.074$\pm$0.090  & 2.814$\pm$0.027 & 182.3$\pm$6.61 & 0.885$\pm$0.044 & 1.059$\pm$0.038 \\
\textit{MultiEmb-Add} & 2.683$\pm$0.043 & 171.0$\pm$11.5 & \textbf{0.782*$\pm$0.006} & 1.070$\pm$0.084  & 2.874$\pm$0.059 & 188.5$\pm$1.36 & 0.846$\pm$0.002 & 1.062$\pm$0.019 \\
\textit{MultiEmb-Gating} & 2.651$\pm$0.037 & \textbf{153.2$\pm$8.51} & 0.784*$\pm$0.006 & \cellcolor{blue!10}1.026*$\pm$0.022 & 2.700*$\pm$0.011 &  201.3$\pm$19.6 & 0.826$\pm$0.007 & 1.057$\pm$0.079 \\
\textit{LSTM-Concat} & 2.589*$\pm$0.007 & 157.0$\pm$5.74 & 0.786*$\pm$0.005 & \cellcolor{blue!10}1.027*$\pm$0.032 & 2.818$\pm$0.022 & 190.5$\pm$3.01 & 0.850$\pm$0.011 & 1.061$\pm$0.017 \\

\textit{LSTM-Add} & \textbf{2.577*$\pm$0.003} & 162.4$\pm$2.07 & 0.783*$\pm$0.004 & \cellcolor{blue!10}1.036*$\pm$0.055  & 2.825$\pm$0.030 & 192.1$\pm$1.63 & 0.837$\pm$0.010 & 1.059$\pm$0.027 \\
\textit{LSTM-Gating} & \textbf{2.577*$\pm$0.010} & 163.2$\pm$8.99 & 0.787*$\pm$0.003 & 1.040$\pm$0.056 & 2.692*$\pm$0.030 & 202.4$\pm$4.53 & 0.816$\pm$0.009 & 1.053$\pm$0.080 \\
\bottomrule
	\end{tabular}
	\end{center}
\vspace{-1em}
\end{table*}

\subsubsection{Late Fusion vs. Early Joint Modeling.} 
In Table \ref{tab: techniquues_60_STMeta_STMGCN_RMSE}, we observe that the \textit{Late Fusion} approaches outperform the \textit{Early Joint Modeling} methods in both the \textit{STMGCN} and \textit{STMeta} backbone networks. Moreover, when compared to the \textit{No Context} baseline, the \textit{EarlyConcat} and \textit{EarlyAdd} techniques exhibit inferior performance. 

As discussed in Section \ref{assumption_discussion}, \textit{Early Joint Modeling} assumes similarities between crowd flow and contextual features. However, this assumption may not always hold. For example, road and crowd flow data only exhibit a moderate positive correlation (i.e., 0.5060, as listed in Table \ref{tab: correlation_score}). These findings suggest that incorporating contextual features via \textit{Early Joint Modeling} may not be the optimal approach, especially if there exist significant disparities between context and crowd flow patterns. Notably, the \textit{Late Fusion} method utilizing gating mechanisms (i.e., \textit{Raw-Gating}) achieves notable performance across various application scenarios and models, indicating their robustness and generalizability. These observations have significant implications for the design of effective spatiotemporal prediction models and can inform future research in this area.

\subsubsection{Is Context Embedding Always Necessary?}
Context embedding is a commonly used technique in previous crowd mobility prediction studies \cite{liu_deeppf:_2019,liang_geoman:_2018,wang_deepsd:_2017,zhang2017deep,yi2019citytraffic}. 
Embedding size is the vital hyper-parameter of embedding layers, we tune their output dimensions and the results are in the Appendix~C. We choose the best embedding size to implement the embedding variants.

From Table \ref{tab: techniquues_60_STMeta_STMGCN_RMSE}, embedding layers enhance model performance by mapping features into a low-dimension space for the methods whose fusion step involves \emph{Concatenate} or \emph{Add}. That is, comparing \emph{Raw-Concat} with \emph{Emb-Concat}/\emph{MultiEmb-Concat}, the latter embedding variants acquire lower \emph{avgNRMSE}. Similar phenomena also occur in the comparison between \emph{Raw-Add} and \emph{Emb-Add}/\emph{MultiEmb-Add}. 
However, in the case of the \emph{Gating} fusion technique, embedding may not be as effective. \emph{Raw-Gating}, which does not use embedding layers, still outperforms \emph{No Context} based on \textit{STMGCN} and even acquires the lowest \emph{avgRMSE} based on \textit{STMeta}. It is worth noting that applying embedding layers before the gated units leads to worse performance on \textit{STMeta}. These findings indicate that embedding layers may work when the fusion methods are \emph{Concatenate} or \emph{Add}. But for \emph{Gating}, embedding layers may not be necessary, and \emph{Raw} features alone can achieve good and generalizable performance.

\subsubsection{Is Past Context Beneficial?} 
Temporal modeling units like \textit{LSTM} can capture time-varying weather features and model their temporal dependencies \cite{ke_short-term_2017}. We evaluate the \textit{LSTM} variants of \textit{Late Fusion} based on \textit{STMGCN} and \textit{STMeta}. The results are in Table \ref{tab: techniquues_60_STMeta_STMGCN_RMSE}. 

It gives us the following insights. First, whether past contextual features perform better may depend on the applications they apply to. In bike and EV, \textit{LSTM} variants consistently outperform \textit{No Context}, while past contextual features lead to worse performance in metro applications. This could be because subway operations are less influenced by past temporal context factors (e.g., weather conditions in the previous few hours). Second, the efficacy of past contextual features depends on the spatiotemporal backbone networks used. For example, \textit{LSTM-Add} gets better prediction than \textit{No Context} when based on \textit{STMeta} but performs worse with \textit{STMGCN}. Third, even though \textit{LSTM} variants show better results than \textit{No Context}, they are still not as good as \textit{Raw-Gating}.
These findings highlight the need to explore more advanced and robust techniques for modeling historical context beyond \textit{LSTM} variants to achieve accurate crowd flow prediction.

\subsubsection{Robustness Analysis under Different Temporal Intervals.}
To evaluate the performance of different modeling methods on tasks with different temporal intervals, we conduct experiments on 30-minute and 120-minute prediction tasks, as detailed in Table \ref{tab: techniquues_30_120_STMeta_RMSE}. Results indicate that the performance of different modeling methods remains consistent across different interval tasks. For instance, \textit{Raw-Gating} and \textit{LSTM-Add} outperform \textit{No Context} in 60-minute tasks, and both are still more effective than \textit{No Context} for 30 and 120-minute tasks. Notably, \textit{Raw-Gating} achieves the best result for 30-minute tasks and ranks second-best for 120-minute tasks, demonstrating its great generalizability across different temporal interval tasks.

\begin{table}[htbp!]
	\footnotesize
	\caption{30/120-minute RMSE results of different modeling techniques based on \textit{STMeta}. The best results are highlighted in bold. \emph{No Context} does not incorporate any contextual features. The modeling techniques with better \textit{avgNRMSE} than \emph{No Context} are highlighted in {\color{blue!70}blue}.}
    \vspace{-2em}
	\label{tab: techniquues_30_120_STMeta_RMSE}
    \renewcommand\tabcolsep{2pt}
	\begin{center}
	\resizebox{0.49\textwidth}{!}{
	
    \begin{tabular}{lccccccccccccccccccccccc}
	\toprule
	& \multicolumn{4}{c}{\textbf{30-minute}} & \multicolumn{4}{c}{\textbf{120-minute}}\\
	\cmidrule(lr){2-5} \cmidrule(lr){6-9}

    \multicolumn{1}{c}{} & \multicolumn{1}{c}{\textit{Bike}} & \multicolumn{1}{c}{\textit{Metro}} & \multicolumn{1}{c}{\textit{EV}} & \multicolumn{1}{c}{\textit{avgNRMSE}} & \multicolumn{1}{c}{\textit{Bike}} & \multicolumn{1}{c}{\textit{Metro}} & \multicolumn{1}{c}{\textit{EV}} & \multicolumn{1}{c}{\textit{avgNRMSE}} \\

\midrule

\textit{No Context} & 2.211 & 77.62 & 0.672 & 1.047 & 3.830 & 339.6 & 0.955 & 1.053 \\ 
\textit{EarlyConcat} & 3.173 & 119.5 & 0.967 & 1.541 & 4.731 & 421.5 & 1.653 & 1.481 \\ 
\textit{EarlyAdd} & 2.485 & 78.99 & 0.718 & 1.121 & 3.991 & 361.1 & 1.322 & 1.227 \\ 
\textit{Raw-Concat} & 2.229 & 83.51 & 0.658 & 1.069 & 3.834 & 544.8 & 0.955 & 1.260 \\ 
\textit{Raw-Add} & 2.205 & 84.63 & 0.676 & 1.080 & 3.797 & 556.3 & 1.045 & 1.302 \\ 
\textit{Raw-Gating} & 2.173 & \textbf{74.40} & \textbf{0.640} & \cellcolor{blue!10}\textbf{1.010} & 3.741 & 334.9 & 0.906 & \cellcolor{blue!10}1.021 \\ 
\textit{Emb-Concat} & 2.199 & 77.35 & 0.662 & \cellcolor{blue!10}1.039 & 3.840 & 375.8 & 0.899 & 1.069 \\
\textit{Emb-Add} & 2.124 & 80.04 & 0.669 & \cellcolor{blue!10}1.043 & 3.794 & 345.3 & 0.902 & \cellcolor{blue!10}1.035 \\ 
\textit{Emb-Gating} & 2.189 & 91.76 & 0.656 & 1.099 & 3.758 & 381.1 & 0.899 & 1.067 \\ 
\textit{MultiEmb-Concat} & 2.133 & 78.91 & 0.670 & \cellcolor{blue!10}1.040 & 3.800 & \textbf{330.6} & 0.929 & \cellcolor{blue!10}1.031 \\ 
\textit{MultiEmb-Add} & 2.254 & 88.05 & 0.665 & 1.097 & 3.776 & 388.1 & 0.914 & 1.081 \\ 
\textit{MultiEmb-Gating} & 2.208 & 85.03 & 0.670 & 1.079 & 3.788 & 371.3 & \textbf{0.884} & 1.054 \\ 
\textit{LSTM-Concat} & 2.116 & 77.23 & 0.665 & \cellcolor{blue!10}1.027 & 3.737 & 350.6 & 0.902 & \cellcolor{blue!10}1.035 \\  
\textit{LSTM-Add} & \textbf{2.109} & 76.96 & 0.656 & \cellcolor{blue!10}1.020 & 3.691 & 343.4 & 0.889 & \cellcolor{blue!10}\textbf{1.019} \\ 
\textit{LSTM-Gating} & 2.167 & 78.99 & 0.657 & \cellcolor{blue!10}1.039 & \textbf{3.648} & 352.4 & 0.893 & \cellcolor{blue!10}1.025 \\ 
\bottomrule
	\end{tabular}}
	\end{center}
    \vspace{-1em}
\end{table}

\subsubsection{Do Context Modeling Techniques Increase the Computational Burden?} 
\begin{figure}[htbp]
	\centering
	\includegraphics[width=0.98\linewidth]{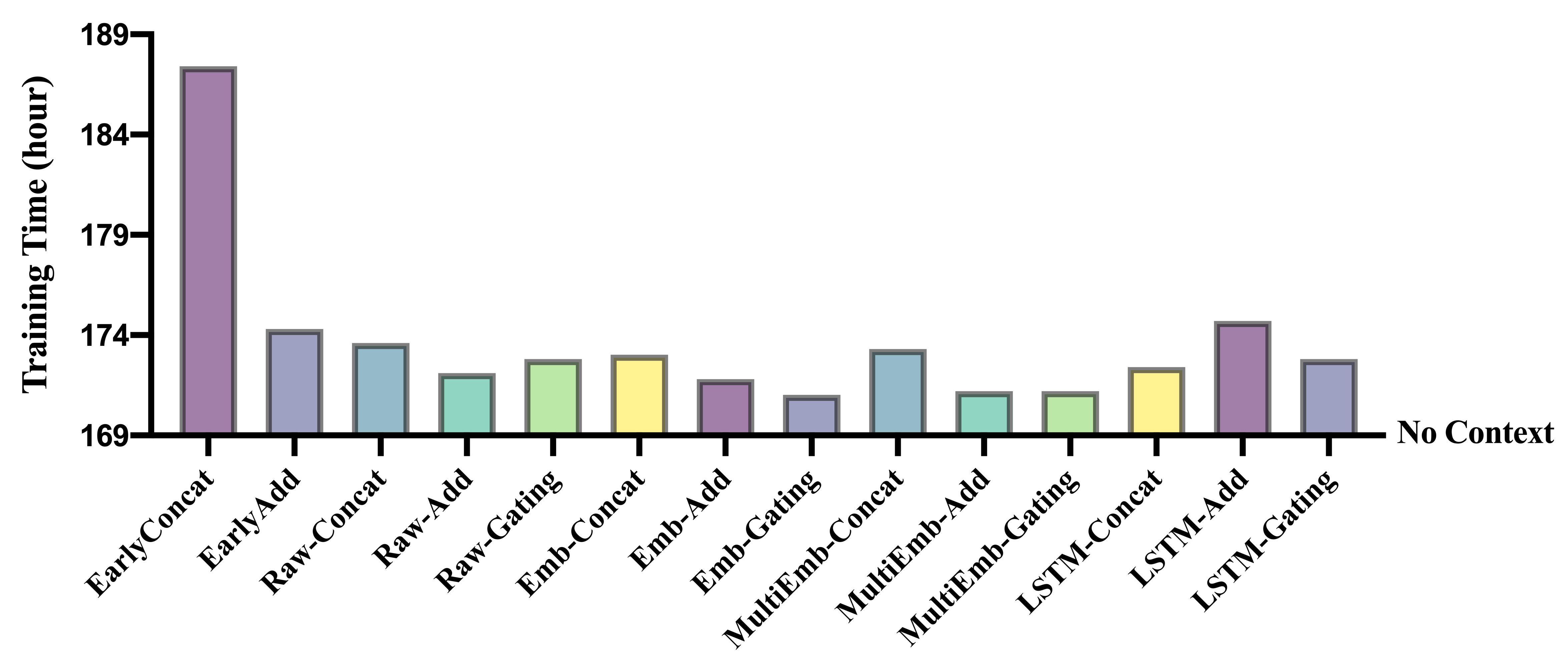}
    \vspace{-1em}
	\caption{The training time of different context modeling techniques based on \textit{STMeta} in the Bike Chicago datasets. \emph{EarlyConcat} is more time-consuming because it incurs more parameters in spatiotemporal modeling units.}
	\label{fig:timeHour}
    \vspace{-1em}
\end{figure}

Intuitively, context modeling techniques increase computational burdens as we need to train more parameters in neural networks. We examine the training hours of different context modeling techniques as illustrated in Figure \ref{fig:timeHour} based on \textit{STMeta} in the Bike Chicago dataset. We observe that most context-engaged models need more training time compared to \emph{No Context}. More specifically, most context modeling techniques only need an additional five hours (about 3\%) for training, while \emph{No Context} needs about 169 hours. This suggests that existing context modeling techniques are less complex than spatiotemporal modeling units. Besides, \emph{EarlyConcat}, which fuses contextual features with raw inputs before applying spatiotemporal modeling, requires around 187 hours for training, the most among the methods discussed. Despite this, it only adds an extra 11\% training time. In conclusion, existing context modeling techniques generally impose slight (usually less than 3\%) computational burdens.

\subsection{Results and Analysis on Contextual Features} 

To study the generalizability of contextual features, we conduct experiments by utilizing various combinations of contextual features, including weather, holiday, temporal position, POIs, road, demographic, and spatial position. We choose \emph{Raw-Gating} as the context modeling technique since it has good generalizability as observed in Table \ref{tab: techniquues_60_STMeta_STMGCN_RMSE}. 
As later shown in our experiments, using more contextual features doesn't always yield better results. With seven types of contextual features resulting in $2^7$ (128) feature combinations, testing them all is not only inefficient but also impractical. Hence, we first incorporate individual contextual features and only combine them with others if we observe significant improvements.
The results are shown in Table \ref{tab: features_60_STMeta_RMSE} and Table \ref{tab: features_30_120_STMeta_RMSE}. Each row is named as the features in consideration. For example, \emph{Wea-Holi} includes weather and holiday features. To our surprise, more kinds of contextual features will not always bring better performance through existing context modeling techniques. Especially, we observe that \textit{Holi-TP} consistently performs better than \textit{No Context} across 30/60/120-minute tasks, demonstrating its generalizability.

\begin{table}[htbp]
	\footnotesize
	\caption{60-minute RMSE results of different contextual features based on \textit{STMeta}. The best results are in bold. The feature combinations marked with * significantly ($p < 0.05$) outperform \emph{No Context}. Those with significantly better \textit{avgNRMSE} than \emph{No Context} are highlighted in {\color{blue!70}blue}. (Wea: Weather; Holi: Holiday; TP: Temporal Position; POIs: Point of Interests; Demo: Demographic; SP: Spatial Position)}
    \vspace{-1.5em}
	\label{tab: features_60_STMeta_RMSE}
    \renewcommand\tabcolsep{2.0pt} 
	\begin{center}
    \resizebox{0.48\textwidth}{!}{
    \begin{tabular}{lcccccccccccccccccccccccccccccccccc}
    \toprule
    \multicolumn{5}{c}{\textbf{STMeta}}\\
    \cmidrule(lr){2-4} \cmidrule(lr){5-5}
    & \textit{Bike} & \textit{Metro} & \textit{EV} & \textit{avgNRMSE} \\
    \midrule

\textit{No Context} & 2.699$\pm$0.036 & 155.9$\pm$11.6 & 0.814$\pm$0.006 & 1.130$\pm$0.135  \\ 
\midrule
\multicolumn{2}{l}{\textbf{Temporal Contextual Feature}} \\
\textit{Wea} & 2.676$\pm$0.010 & 171.6$\pm$2.75 & 0.813$\pm$0.006 & 1.170$\pm$0.213 \\ 
\textit{Holi} & 2.636*$\pm$0.013 & 154.0$\pm$1.92 & 0.783*$\pm$0.002 & \cellcolor{blue!10}1.103*$\pm$0.115  \\ 
\textit{TP} & 2.577*$\pm$0.010 & 124.6*$\pm$7.47 & 0.798*$\pm$0.004 & \cellcolor{blue!10}1.021*$\pm$0.016  \\ 

\textit{Wea-Holi} & 2.681$\pm$0.026 & 166.9$\pm$2.62 & 0.797*$\pm$0.004 & 1.151$\pm$0.196  \\ 
\textit{Wea-TP} & 2.580*$\pm$0.027 & 150.4$\pm$6.28 & 0.787*$\pm$0.008 & 1.088$\pm$0.133  \\ 
\textit{Holi-TP} & \textbf{2.555*$\pm$0.004} & \textbf{124.2*$\pm$0.12} & 0.776*$\pm$0.004 & \cellcolor{blue!10} \textbf{1.008*$\pm$0.014}  \\ 
\textit{Wea-Holi-TP} & 2.612*$\pm$0.014 & 153.4$\pm$7.31 & 0.785*$\pm$0.004 & 1.100$\pm$0.143  \\ 
\midrule
\multicolumn{2}{l}{\textbf{Spatial Contextual Feature}} \\
\textit{POIs} & 2.674$\pm$0.003 & 149.8$\pm$3.66 & 0.799*$\pm$0.001 & \cellcolor{blue!10} 1.104*$\pm$0.114  \\ 
\textit{Road} & 2.717$\pm$0.029 & 156.8$\pm$8.80 & 0.801*$\pm$0.004 & 1.131$\pm$0.140 \\ 
\textit{Demo} & 2.727$\pm$0.019 & 168.1$\pm$2.98 & 0.804$\pm$0.005 & 1.164$\pm$0.192 \\ 
\textit{SP} & 2.690$\pm$0.027 & 169.6$\pm$3.61 & 0.793*$\pm$0.006 & 1.159$\pm$0.207 \\ 

\midrule
\multicolumn{5}{l}{\textbf{Temporal Contextual Feature \& Effective Spatial Contextual Feature}} \\
\textit{Wea-POIs} & 2.695$\pm$0.041 & 175.8$\pm$16.0 & 0.813$\pm$0.007 & 1.183$\pm$0.228  \\ 
\textit{Holi-POIs} & 2.662$\pm$0.030 & 153.6$\pm$4.74 & 0.786*$\pm$0.003 & 1.107$\pm$0.139   \\ 
\textit{TP-POIs} & 2.587*$\pm$0.015 & 125.7*$\pm$10.3 & 0.786*$\pm$0.001 & \cellcolor{blue!10} 1.020*$\pm$0.022  \\ 

\textit{Wea-Holi-POIs} & 2.668$\pm$0.041 & 172.3$\pm$12.0 & 0.800$\pm$0.011 & 1.166$\pm$0.223  \\ 
\textit{Wea-TP-POIs} & 2.605*$\pm$0.015 & 164.6$\pm$7.39 & 0.791*$\pm$0.004 & 1.132$\pm$0.195 \\ 
\textit{Holi-TP-POIs} & 2.603*$\pm$0.047 & 131.4*$\pm$5.90 & 0.778*$\pm$0.003 & \cellcolor{blue!10} 1.034*$\pm$0.044  \\ 
\textit{Wea-Holi-TP-POIs} & 2.599*$\pm$0.002 & 149.9$\pm$9.57 & 0.780*$\pm$0.001 & 1.087$\pm$0.130 \\  

\midrule
\multicolumn{5}{l}{\textbf{Spatial Contextual Feature \& Effective Temporal Contextual Feature}} \\
\textit{Holi-TP-POIs} & 2.603*$\pm$0.047 & 131.4*$\pm$5.90 & 0.778*$\pm$0.003 & \cellcolor{blue!10} 1.034*$\pm$0.044  \\ 
\textit{Holi-TP-Road} & 2.579*$\pm$0.015 & 136.0*$\pm$10.1 & 0.778*$\pm$0.002 & \cellcolor{blue!10} 1.047*$\pm$0.067 \\ 
\textit{Holi-TP-Demo} & 2.609*$\pm$0.012 & 146.0$\pm$9.09 & 0.773*$\pm$0.004 & \cellcolor{blue!10} 1.075*$\pm$0.112 \\ 
\textit{Holi-TP-SP} & 2.640$\pm$0.088 & 149.6$\pm$7.42 & \textbf{0.769*$\pm$0.002} & \cellcolor{blue!10} 1.087*$\pm$0.128 \\ 
\midrule
\textit{All} & 2.626*$\pm$0.020 & 180.9$\pm$15.3 & 0.773*$\pm$0.007 & 1.173$\pm$0.275 \\  
\bottomrule
\end{tabular}}
	\end{center}
\end{table}

\subsubsection{Temporal Context vs. Spatial Context.} 
In Table \ref{tab: features_60_STMeta_RMSE}, we observe that weather features improve bike application performance but not in metro and EV applications, indicating weather's limited generalizability across different applications. It also suggests that weather may have a great impact on bike applications, which is consistent with our daily routines (e.g., rain affects bicycle travel) \cite{li_traffic_2015,hoang_fccf:_2016,wuInterpreting2016}. Additionally, \textit{Holi} and \textit{TP} generally enhance prediction performance, significantly outperforming \textit{No Context}. Notably, \textit{Holi-TP}, which combines holiday and temporal position, consistently reduces RMSE by over 4.7\%, demonstrating strong generalizability.

For spatial context, only \textit{POIs} shows significant improvement compared to \textit{No Context}. However, in comparison to \textit{Holi-TP}, \textit{POIs} only slightly enhance performance (with most improvements being less than 2\%). Besides, \textit{Demo}, \textit{Road}, and \textit{SP} do not significantly improve predictions and may even introduce noise that decreases the performance.
In summary, the above findings highlight the advantages of temporal contextual features, particularly the combination of holiday and temporal position, which provide more generalizable and beneficial information.

\subsubsection{Do More Contextual Features Always Result in Better Prediction?} Intuitively, we may hold a wide belief that the more contextual features we use, the better prediction we get. However, this hypothesis has been challenged by our findings, which reveal that this is not always the case.
Specifically, in Table~\ref{tab: features_60_STMeta_RMSE}, Table~\ref{tab: features_30_120_STMeta_RMSE}, and Table~18 (available in the online appendix), our results indicate that across various machine learning models, the optimal average performance (\emph{avgRMSE}) is attained only when two features, namely holiday and temporal position (\emph{Holi-TP}), are considered. The uncovered insight is unexpected, particularly given that conventional features such as weather have not exhibited significant improvement in generalizability, as revealed by our results. The potential factors contributing to this phenomenon include the lack of adequate granularity in the selected features - for example, weather patterns tend to be uniform across different locations within a city. Alternatively, the relatively simplistic nature of current context modeling methods may be insufficient to capture the intricate patterns involved in complex features such as weather. Consequently, further research examining effective methods of incorporating contextual features into spatiotemporal prediction models is imperative.

\subsubsection{Do Contextual Features Perform Consistently in Different Interval Prediction Tasks?} 

We evaluate contextual feature effectiveness in prediction tasks with 30-minute and 120-minute temporal intervals. The results are in Table~\ref{tab: features_30_120_STMeta_RMSE}. 
 
\begin{table}[htbp]
	\footnotesize
	\caption{30/120-minute RMSE results of different contextual features based on \textit{STMeta}. The best results are in bold. The feature combinations with better \textit{avgNRMSE} than \emph{No Context} are highlighted in {\color{blue!70}blue}.}
    \vspace{-2em}
	\label{tab: features_30_120_STMeta_RMSE}
	\renewcommand\tabcolsep{2pt}
	\begin{center}
    \resizebox{0.49\textwidth}{!}{
	\begin{tabular}{lccccccccccccccccccccccccccc}
\toprule

\multirow{2}{*}{\textbf{STMeta}} & \multicolumn{4}{c}{\textbf{30-minute}} & \multicolumn{4}{c}{\textbf{120-minute}} \\

\cmidrule(lr){2-5} \cmidrule(lr){6-9} 

& Bike & Metro & EV & \textit{avgNRMSE} & Bike & Metro & EV & \textit{avgNRMSE} \\
\midrule
\textit{No Context} & 2.211 & 77.62 & 0.672 & 1.071 & 3.830 & 339.6 & 0.955 & 1.081 \\ 
\midrule
\textit{Wea} & 2.355 & 83.31 & 0.697 & 1.134 & 3.792 & 394.5 & 0.947 & 1.132 \\ 
\textit{Holi} & 2.118 & 77.16 & 0.650 & \cellcolor{blue!10}1.042 & 3.524 & 338.5 & 0.901 & \cellcolor{blue!10}1.030 \\ 
\textit{TP} & \textbf{2.065} & \textbf{71.15} & \textbf{0.639} & \cellcolor{blue!10}\textbf{1.000} & 3.547 & 320.4 & 0.922 & \cellcolor{blue!10}1.021 \\ 
\textit{Wea-Holi} & 2.271 & 81.50 & 0.652 & 1.089 & 3.780 & 351.0 & 0.957 & 1.089 \\ 

\textit{Wea-TP} & 2.183 & 76.27 & 0.694 & 1.072 & 3.647 & 333.5 & 0.931 & \cellcolor{blue!10}1.048 \\ 

\textit{Holi-TP} & 2.109 & 72.85 & 0.646 & \cellcolor{blue!10}1.019 & \textbf{3.500} & \textbf{315.9} & 0.897 & \cellcolor{blue!10}\textbf{1.003} \\ 

\textit{Wea-Holi-TP} & 2.173 & 77.40 & 0.640 & \cellcolor{blue!10}1.047 & 3.741 & 336.9 & 0.906 & \cellcolor{blue!10}1.051 \\ 
\midrule
\textit{POIs} & 2.130 & 77.90 & 0.693 & \cellcolor{blue!10}1.070 & 3.922 & 332.4 & 0.942 & \cellcolor{blue!10}1.077 \\
\textit{Road} & 2.122 & 76.34 & 0.667 & \cellcolor{blue!10}1.048 & 3.887 & 327.7 & 1.017 & 1.097\\
\textit{Demo} & 2.098 & 73.41 & 0.667 & \cellcolor{blue!10}1.031 & 3.764 & 339.1 & 0.971 & \cellcolor{blue!10}1.080\\
\textit{SP} & 2.187 & 76.36 & 0.659 & \cellcolor{blue!10}1.055 & 3.827 & 360.8 & 0.966 & 1.107\\
\midrule
\textit{Wea-POIs} & 2.164 & 91.39 & 0.732 & 1.159 & 3.788 & 362.1 & 0.960 & 1.102 \\ 
\textit{Holi-POIs} & 2.126 & 76.96 & 0.669 & \cellcolor{blue!10}1.053 & 3.893 & 342.7 & 0.914 & \cellcolor{blue!10}1.075 \\ 
\textit{TP-POIs} & 2.085 & 71.74 & 0.669 & \cellcolor{blue!10}1.022 & 3.836 & 328.8 & 0.943 & \cellcolor{blue!10}1.065 \\ 
\textit{Wea-Holi-POIs} & 2.170 & 83.93 & 0.669 & 1.092 & 3.729 & 350.3 & 0.962 & 1.085 \\ 
\textit{Wea-TP-POIs} & 2.160 & 84.17 & 0.671 & 1.093 & 3.872 & 348.6 & 0.932 & 1.086 \\ 
\textit{Holi-TP-POIs} & 2.154 & 76.41 & 0.670 & \cellcolor{blue!10}1.055 & 3.824 & 321.2 & 0.921 & \cellcolor{blue!10}1.048 \\ 
\textit{Wea-Holi-TP-POIs} & 2.106 & 82.38 & 0.680 & 1.081 & 3.858 & 340.6 & 0.924 & \cellcolor{blue!10}1.073 \\ 
\midrule
\textit{Holi-TP-POIs} & 2.154 & 76.41 & 0.670 & \cellcolor{blue!10}1.052 & 3.824 & 321.2 & 0.921 & \cellcolor{blue!10}1.045 \\ 
\textit{Holi-TP-Road} & 2.101 & 75.36 & 0.659 & \cellcolor{blue!10}1.036 & 3.633 & 333.8 & 0.922 & \cellcolor{blue!10}1.044\\ 
\textit{Holi-TP-Demo} & 2.130 & 72.76 & 0.651 & \cellcolor{blue!10}1.024 & 3.744 & 332.5 & \textbf{0.890} & \cellcolor{blue!10}1.041\\
\textit{Holi-TP-SP} & 2.123 & 73.39 & 0.646 & \cellcolor{blue!10}1.024 & 3.749 & 344.1 & 0.895 & \cellcolor{blue!10}1.055 \\  
\midrule
\textit{All} & 2.129 & 80.75 & 0.650 & \cellcolor{blue!10}1.061 & 3.724 & 341.6 & 0.894 & \cellcolor{blue!10}1.050\\
\bottomrule
		\end{tabular}}
	\end{center}
\vspace{-1em}
\end{table}

Weather features do not improve the overall performance compared to \textit{No Context}. \textit{Holi} and \textit{TP} still enhance prediction performance as they perform in 60-minute tasks. Notably, temporal position features were more effective in 30-minute prediction tasks, providing more fine-grained information. Spatial features in 30- and 120-minute tasks perform consistently with the 60-minute tasks, with slightly better or worse improvement, likely due to their relative stability over long periods. In summary, holiday features show benefits across varying temporal intervals, and temporal position features proved more effective in short-interval prediction tasks (e.g., 30 minutes).

\section{Guidelines and Their Generalizability}
\subsection{Guidelines and Insights} \label{sec: guidelines}
In general, based on our benchmark, we find that adding contextual features may not always increase STCFP prediction accuracy, and practitioners should carefully determine which features and modeling techniques are applied.

\subsubsection{Contextual Features Guidelines} (i) Weather, holiday, temporal position, POIs, road, demographic, and spatial position could be considered in almost all kinds of STCFP scenarios. However, using more contextual features may not always result in better predictions. (ii) The feature combination of holiday and temporal position can provide more generalizable beneficial information. Additionally, temporal position and holiday features are easy to access and thus are prior recommended. (iii) Temporal position features are especially critical for short-interval prediction tasks.

\subsubsection{Modeling Techniques Guidelines} (i) Existing early joint modeling techniques (\emph{EarlyConcat} and \emph{EarlyAdd}) are not good enough to extract beneficial information from contextual features. That is to say, fusing context in the high-level layer of neural networks is a better choice rather than in the low-level layer. (ii) In the late fusion techniques, the embedding layers may work when the fusion methods are \emph{Concat} and \emph{Add}. But for \emph{Gating}, embedding layers may not be necessary. (iii) \emph{Gating} is highly recommended for context fusion because it has the best generalizability across three scenarios and two state-of-the-art spatiotemporal neural network models. (iv) We recommend \emph{Raw-Gating} as the default technique to try when building a new spatiotemporal prediction neural network model. It not only has a good and generalizable performance across different experiment scenarios but also does not need to tune many fusing parameters (e.g., embedding size).

\subsection{Generalize to New Applications}
In Section~\ref{sec: guidelines}, we conclude two guidelines: (i) the combination of holiday and temporal position is generally beneficial; (ii) \emph{Raw-Gating} is more generalizable than other investigated context modeling techniques. To test the generalizability of these two guidelines, we conduct experiments on three new applications, namely the bike demand prediction tasks in NYC (bike-sharing), the traffic speed prediction task in California, and the pedestrian count task in Melbourne. Dataset details are in Appendix~A.

\begin{table}[htbp]
	\small
	\caption{Results on three new applications (i.e., Bike NYC, PEMS BAY, and Pedestrian Count), with the inclusion of holiday and temporal position features (Holi-TP). This experiment is conducted based on \textit{STMeta}.}
    \vspace{-1.5em}
	\label{tab: new_applications}
	\begin{center}
    \resizebox{0.48\textwidth}{!}{
    \begin{tabular}{lcccccccccc}
    \toprule
    
     & \multicolumn{2}{c}{\textbf{Bike NYC}} & \multicolumn{2}{c}{\textbf{PEMS BAY}} & \multicolumn{2}{c}{\textbf{Pedestrian Count}}\\
    \cmidrule(lr){2-3} \cmidrule(lr){4-5} \cmidrule(lr){6-7} 
    & \textit{RMSE} & \textit{MAE} & \textit{RMSE} & \textit{MAE} & \textit{RMSE} & \textit{MAE}  \\
    \midrule
\emph{No Context} & 3.569 & 2.189 & 3.297 & 1.581 & 107.7 & 52.83 \\ 
\emph{Raw-Gating} & 3.395 & 2.102 & 3.253 & 1.555 & 106.5 & 50.93 \\ 
\midrule
\emph{Improvement} & \textbf{+4.88\%} & \textbf{+3.94\%} & \textbf{+1.33\%} & \textbf{+1.62\%} & \textbf{+1.11\%} & \textbf{+3.60\%} \\
    \bottomrule
    \end{tabular}}
	\end{center}
\end{table}

In Table~\ref{tab: new_applications}, \textit{Raw-Gating (Holi-TP)} incorporates the combination of holiday and temporal position by \emph{Raw-Gating}. For comparison, \textit{No Context} shows the results that no contextual features are taken in. The experiments are conducted based on \textit{STMeta}. We observe that, without meticulously selecting the context modeling technique and contextual feature combinations, we directly incorporate holiday and temporal position by \emph{Raw-Gating} and get remarkable improvement compared to the model that doesn't consider any contextual features. With contextual features, the prediction performance of these three applications consistently improves (up to 4.88\% and 3.94\% in terms of RMSE and MAE, respectively), demonstrating the generalizability of our guidelines. However, our guidelines don't mean the combination of holiday and temporal position and \textit{Raw-Gating} are always capable for every STCFP application. Besides, even though we may get better performance, we still may benefit from more robust and advanced context modeling techniques or taking other contextual feature combinations.

\subsection{Generalize to New ST Models}
The main guidelines for taking context into spatiotemporal models, namely the incorporation of holiday and temporal position through the \textit{Raw-Gating} technique, were developed based on experiments using \textit{STMGCN} and \textit{STMeta}. However, it is important to investigate whether these guidelines apply to other state-of-the-art spatiotemporal models. To address this question, we extend the analysis to include two additional models, namely \textit{GraphWaveNet}~\cite{graphwavenet_2019} and \textit{AGCRN}~\cite{AGCRN_2020}, both of which have demonstrated efficacy in prior studies~\cite{dynamic_graph_benchmark_2023}. The experiments were conducted on the Bike NYC dataset, and the results are reported in Table~\ref{tab: new_stModel}.

\begin{table}[htbp]
	\small
	\caption{Results on two new spatiotemporal models, \textit{GraphWaveNet} and \textit{AGCRN}, with the inclusion of holiday and temporal position features (Holi-TP). The experiment is performed on the Bike NYC dataset.}
    \vspace{-1.5em}
	\label{tab: new_stModel}
	\begin{center}
    \begin{tabular}{lcccccccccc}
\toprule
& \multicolumn{2}{c}{\textbf{GraphWaveNet}} & \multicolumn{2}{c}{\textbf{AGCRN}} \\
\cmidrule(lr){2-3} \cmidrule(lr){4-5} 
& \textit{RMSE} & \textit{MAE} & \textit{RMSE} & \textit{MAE} \\
\midrule
\emph{No Context} & 3.466 & 2.085 & 3.620 & 2.384 \\ 
\emph{Raw-Gating} & 3.401 & 2.037 & 3.564 & 2.339 \\ 
\midrule
\emph{Improvement}  & \textbf{+1.88\%} & \textbf{+2.30\%} & \textbf{+1.55\%} & \textbf{+1.89\%} \\
    \bottomrule
    \end{tabular}
	\end{center}
\end{table}

The results in Table~\ref{tab: new_stModel} demonstrate that various spatiotemporal models, including \textit{GraphWaveNet}, \textit{AGCRN}, and \textit{STMeta}, derive benefits from the incorporation of contextual features using the \emph{Raw-Gating} technique. These findings highlight the importance of context modeling techniques in providing an alternative perspective for characterizing crowd flow dynamics. Notably, such techniques offer complementary insights to spatiotemporal modeling approaches, and their development is likely to have broader implications for enhancing the performance of spatiotemporal models in a range of settings.

\section{Lightweight Model-based Contextual Feature Selection Strategy}

Although \textit{Holi-TP} exhibits commendable generalizability, the most effective combinations of context features differ across datasets. For example, as shown in Table \ref{tab: MGateConcat_RMSE}, \textit{Holi-TP} achieves superior performance in the Bike and Metro datasets, whereas \textit{Holi-TP-SP} excels in the EV dataset. Therefore, it is non-trivial to select the most effective feature combination for specific applications.

\begin{table}[htbp]
	\footnotesize
	\caption{RMSE results of \textit{Holi-TP} and \textit{Holi-TP-SP} based on \textit{STMeta}. It shows that the most effective combinations of context features vary across different datasets. The best results are in bold. (Holi: Holiday; TP: Temporal Position; SP: Spatial Position)}
    \vspace{-2em}
	\label{tab: MGateConcat_RMSE}
	\begin{center}
	\resizebox{0.49\textwidth}{!}{
    \begin{tabular}{lccccccccccccccccccccccccccccccccc}
	\toprule
			
	\textbf{STMeta} & \textit{Bike} &\textit{Metro} & \textit{EV} \\
	\midrule
\textit{No Context} & 2.699$\pm$0.036 & 155.9$\pm$11.6 & 0.814$\pm$0.006  \\
\midrule
\textit{Raw-Gating (Holi-TP)} & \textbf{2.555$\pm$0.004} & \textbf{124.2$\pm$0.12} & 0.776$\pm$0.004 \\
\textit{Raw-Gating (Holi-TP-SP)} & 2.640$\pm$0.088 & 149.6$\pm$7.42 & \textbf{0.769$\pm$0.002} \\ 
	\bottomrule
	\end{tabular}}
	\end{center}
\end{table}

One approach to discovering effective feature combinations is by training separate deep models for each combination and selecting the best-performing one. However, this strategy is impractical due to the time-consuming nature of training deep models. We rank different feature combinations using \textit{XGBoost} and \textit{STMeta} in Table \ref{tab: result_rank} according to the benchmark results of \textit{STMeta} (Table \ref{tab: features_60_STMeta_RMSE}) and \textit{XGBoost} (details in Appendix~F). Notably, we find that the effectiveness of context features shows general consistency across models. To confirm this further, we compute the Spearman's rank coefficient between \textit{XGBoost} and \textit{STMeta}, yielding a value of 0.9349, indicating a highly positive monotonic relationship. This suggests that features advantageous to \textit{XGBoost} are likely beneficial for deep models as well. Notably, statistical models like \textit{XGBoost} train much faster than deep models. For instance, on the Bike NYC dataset with the largest number of stations, \textit{XGBoost} takes less than 2 minutes, while \textit{STMeta} requires 169 hours.

\begin{table}[htbp]
	\small
	\caption{The result ranks using different feature combinations based on \textit{XGBoost} and \textit{STMeta}, exhibiting similar trends in their rankings.}
    \vspace{-1.5em}
	\label{tab: result_rank}
	\renewcommand\tabcolsep{15.0pt} 
	\begin{center}
    \begin{tabular}{lcccccccccc}
\toprule
& \textbf{XGBoost} & \textbf{STMeta} \\
\midrule
\emph{Top-1} & \textit{Holi-TP} & \textit{Holi-TP}  \\ 
\emph{Top-2} & \textit{TP-POIs} & \textit{TP-POIs} \\ 
\emph{Top-3} & \textit{TP} & \textit{TP}  \\ 
\emph{Top-4} & \textit{Wea-Holi-TP} & \textit{Holi-TP-POIs}  \\ 
\emph{Top-5} & \textit{Wea-TP} & \textit{Wea-TP} \\
    \bottomrule
    \end{tabular}
	\end{center}
    \vspace{-1.5em}
\end{table}

\begin{figure}[htbp]
	\centering
	\includegraphics[width=0.9\linewidth]{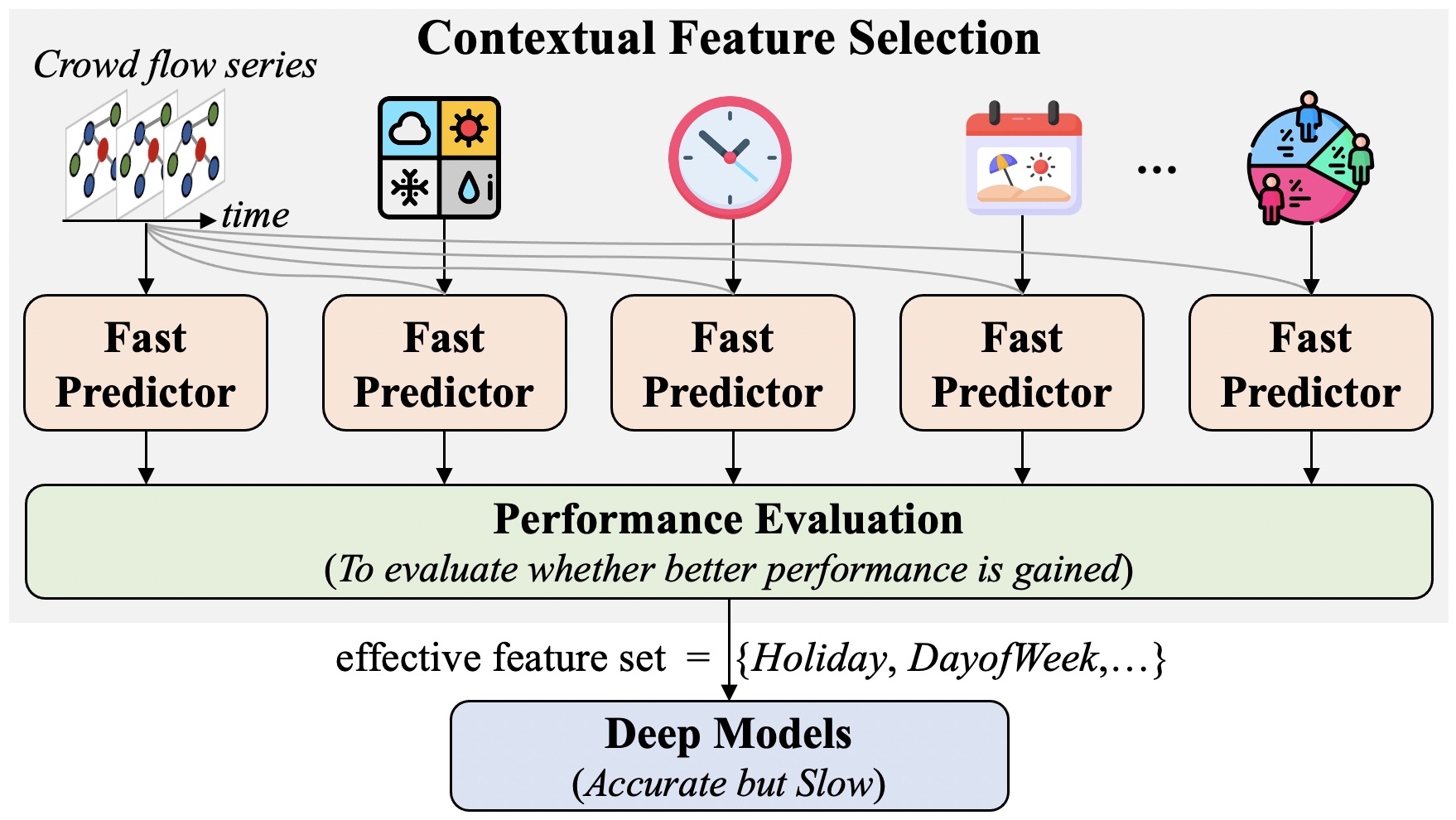}
	\caption{The proposed lightweight model-based contextual feature selection strategy.}
	\label{fig: feature_selection}
\end{figure}

Inspired by this insight, we propose a lightweight model-based contextual feature selection strategy for swiftly identifying effective features, as shown in Figure~\ref{fig: feature_selection}. Our approach suggests evaluating the effectiveness of each feature combination using a 'fast predictor' (e.g., \textit{XGBoost}) and selecting the combination with the lowest prediction error. Subsequently, the most effective feature combination is employed as input for deep models to achieve accurate predictions.

\section{Related Work}
\subsection{Time Series Prediction}
Earlier research regarded crowd mobility prediction as a classic time series prediction problem, assuming that future crowd mobility is linearly correlated with historical observation and using linear models like ARIMA \cite{1995Short}. However, this assumption does not align with the actual characteristics of crowd mobility.
Various nonlinear algorithms have been proposed, including support vector machine \cite{cong2016traffic}, Markov random field \cite{hoang_fccf:_2016}, decision tree method \cite{li_traffic_2015}, and bayesian network \cite{wang2014new}. With the bloom of deep learning techniques, recurrent neural networks (e.g., LSTM \cite{hochreiter1997long} and GRU \cite{chung2014empirical}) and temporal convolutional networks (e.g., Wavenet \cite{WaveNet_2016} and TCN \cite{gated_tcn_2017}) have been proposed to better capture temporal dependencies. Recently, several Transformer variants have emerged for time series prediction, including STTN \cite{xu2020spatial}, Informer \cite{informer_2021}, Autoformer \cite{wu2021autoformer}, FEDformer \cite{FEDformer_2022}, and Non-stationary Transformer \cite{liu2022non}. Although these methods may excel at modeling temporal dependencies, they often overlook spatial correlations and contextual information.

\subsection{Spatio-Temporal Prediction} \label{STCFPreview}
With the enhancement of computing performance and the development of deep learning technology, Convolutional Neural Network (CNN) \cite{ke_short-term_2017,curbGan_2020}, Graph Neural Network (GNN) \cite{chai_multi_graph_2018, yu2018spatio, graphwavenet_2019, song2020spatial, dynamic_multi_traffic_2021, modeling_network_flow_2022} and attention-based methods \cite{ASTGCN_2019, GMAN_AAAI2020, dynamic_attention_2020,li_semantic_gat_2022} are adopted to model spatial dependency in crowd flow prediction problems.
Zhang et al. \cite{zhang2017deep} split the city traffic flow into grids according to time order and then use CNN to capture the spatial dependency. Geng et al. \cite{geng2019spatiotemporal} use a multi-graph convolution model to capture varieties of spatial knowledge. Wang et al. \cite{wang_STDM_survey} give a comprehensive review of recent progress for spatiotemporal prediction from the perspective of spatiotemporal data mining. Jiang et al. \cite{dl_traffic_2021} revisit the deep STCFP models and build a standard benchmark upon popular datasets. Li et al. \cite{dynamic_graph_benchmark_2023} design a dynamic graph convolution recurrent network and propose a benchmark of recent deep STCFP models. Wange et al. \cite{STMeta} design a meta-modeling framework and provide an evaluation benchmark of the STCFP models as well as spatiotemporal knowledge. However, whereas these previous studies attempt to make full use of spatiotemporal correlations and build large-scale benchmarks \cite{STMeta,dynamic_graph_benchmark_2023}, this paper focuses on benchmarking the contextual features and their modeling techniques.

\subsection{Urban Cross-domain Data Fusion} \label{data_fusion_related_work} 
Cross-domain data fusion is crucial for urban computing as it enables a comprehensive understanding of complex urban systems by integrating diverse datasets, including social media, traffic, environment, and geographical data \cite{zheng_data_fusion_2015, zou_deep_fusion_2024}. Deep learning has become dominant in processing and fusing spatiotemporal data in urban computing \cite{wang_STDM_survey}. Recent surveys have delved into the use of data fusion methods to tackle intricate urban challenges. Zheng et al. \cite{zheng_data_fusion_2015} provided a comprehensive overview of data fusion methodologies, highlighting semantic meaning-based data fusion and presenting real-world data fusion examples. On the other hand, Liu et al. \cite{liu_deep_data_fusion_2020} focused on deep learning-based urban big data fusion, categorizing methods into DL-output-based, DL-input-based, and DL-double-stage-based fusion. Zou et al. \cite{zou_deep_fusion_2024} introduce a taxonomy of deep learning for multi-source and multi-modal data fusion in urban computing, categorizing the data perspective, fusion methodology, applications, and exploring the synergy between large language models and urban computing.

Our taxonomy of context modeling techniques builds on the taxonomy introduced in pioneering papers. For example, our late fusion category is similar to the DL-output-based fusion \cite{liu_deep_data_fusion_2020}. However, our work differs in three key aspects. Firstly, we focus on cross-domain data fusion methods for spatiotemporal crowd mobility prediction, allowing for a more comprehensive examination of existing techniques \cite{ke_short-term_2017,zhang_flow_2019, lin2019deepstn+} (e.g., 14 techniques are examined). Secondly, we analyze the assumptions underlying these modeling techniques and offer a detailed analysis that offers several guidelines that inspire researchers and practitioners. Thirdly, we introduce a practical framework for selecting contextual features to alleviate performance degradation.

\section{Conclusions and Future Work} 
In this paper, we explore the generalizability of contextual features and context modeling techniques for crowd mobility prediction. We conduct analytical and experimental studies. In the analytical studies, we investigate contextual features and modeling techniques in the literature, developing a comprehensive taxonomy of context modeling techniques. In the experimental benchmark, we analyze the generalizability of various contextual features (weather, holiday, temporal position, POIs, road, demographic, and spatial position) and diverse modeling techniques using state-of-the-art spatiotemporal prediction models (i.e., \textit{STMGCN} and \textit{STMeta}). We create an experimental platform with large-scale crowd mobility data, contextual data, and prediction models. Additionally, we provide suggestions for practitioners interested in incorporating contextual factors into crowd mobility prediction applications.

Our future work would include (1) Advanced modeling techniques. \emph{Raw-Gating}, with good generalizability, is a remarkable technique to model contextual features. We suggest leveraging \emph{Raw-Gating} as the default modeling technique. However, it may still worsen the performance compared to \emph{No Context} in some (although rarely) specific tasks (e.g., 60-minute metro flow prediction based on \textit{STMGCN}). In other words, we have not found a single context modeling technique that can improve the STCFP performance in \textit{every one} of our experimental scenarios. Hence, an advanced context modeling technique is still urgently demanded to benefit the STCFP research community. (2) More datasets and models beyond crowd flow prediction. In addition to urban crowd flow, many other applications are highly dependent on contextual factors. Hence, we plan to extend our benchmark scenarios to a broader scope of spatiotemporal prediction scenarios such as congestion.

\bibliographystyle{IEEEtran}
\bibliography{ref}

\clearpage

\appendices

\section{Dataset Description} \label{dataset_description}
We collect six spatiotemporal crowd flow datasets and corresponding contextual data. Their statistics are in Table \ref{datasetStatistics}. 

\textbf{\textit{Bike-sharing (Bike Chicago and Bike NYC)}.}
The Bike Chicago dataset is collected from Chicago open data portals\footnote{https://www.divvybikes.com/system-data} while the Bike NYC dataset is collected from New York City’s Citi Bike bicycle sharing service\footnote{https://www.citibikenyc.com/system-data}. These datasets are open to everyone for non-commercial purposes, covering a time span of more than one year. Each piece of valid record contains the start station, start time, stop station, stop time, etc. We predict the number of bike-sharing demands in the next hour for each station.

\textbf{\textit{Metro}.}
The metro dataset contains metro trip records in Shanghai, which are obtained by Shanghai Open Data Apps (SODA) challenge\footnote{http://soda.data.sh.gov.cn/index.html}. The time span is three months. Each metro trip record has the check-in time, check-in station, check-out time, and check-out station. We predict the check-in flow amount for all the metro stations.

\textbf{\textit{Electrical Vehicle (EV)}.}
The electrical vehicle (EV) dataset is collected from one major EV charging station operator in Beijing. The time span of the dataset is six months. This dataset contains the occupation situation at different time slots. Each record contains sensing time, available and occupied docks. We predict the number of docks in use for each station as it is the most important demand indicator of the charging stations.

\textbf{\textit{Speed}.} The PEMS BAY dataset records the traffic speed information from 325 sensors in the Bay Area \cite{li2017diffusion}. The time span of the dataset is six months. We predict the traffic speed in the next hour for each sensor.

\textbf{\textit{Pedestrian}.} The Pedestrian Melbourne dataset is collected from the open data website of Melbourne\footnote{https://data.melbourne.vic.gov.au/explore/dataset/pedestrian-counting-system-monthly-counts-per-hour/information/}. We select 55 sensors that record hourly pedestrians count from 2021-01-01 to 2022-11-01. We predict the crowd counts in the next hour for each sensor.

\textbf{\textit{Contextual Features}.}
The weather data, obtained from OpenWeatherMap\footnote{https://openweathermap.org/history-bulk}, provides hourly measurements. We select weather features from the first hour for 120-minute interval tasks. For 30-minute interval tasks, we upsample data by duplicating hourly measurements. We get citywide weather using data from a single meteorological station. Holiday information is parsed using the chinese\_calendar\footnote{https://pypi.org/project/chinesecalendar/0.0.4} and workalendar\footnote{https://github.com/peopledoc/workalendar} packages. Temporal position data is encoded using one-hot encoding for \emph{DayofWeek} and \emph{HourofDay} features. We collect POI data for Shanghai and Beijing via developer APIs from an online map\footnote{https://map.baidu.com}, and for Chicago from OpenStreetMap\footnote{https://www.openstreetmap.org/}. Road data, primarily motorways, trunk, primary, secondary, tertiary, and residential roads, is collected from OpenStreetMap. The demographic data for Chicago, Shanghai, and Beijing are collected from the American Community Survey data website\footnote{https://www.census.gov/programs-surveys/acs}, the Shanghai statistical yearbook\footnote{https://tjj.sh.gov.cn/tjnj/tjnj2016.htm}, and the Beijing statistical yearbook\footnote{https://nj.tjj.beijing.gov.cn/nj/main/2018-tjnj/zk/e/indexce.htm}, respectively. We focus on population density, education, and income statistics. Spatial position data is encoded using one-hot encoding for \emph{CoordofCity} features.

\begin{table}[h]
	\small
    \setcounter{table}{11}
	\caption{Datasets statistics.}
    \vspace{-1em}
	\label{datasetStatistics}
	\begin{center}
		\resizebox{0.49\textwidth}{!}{
\begin{tabular}{lccccccc}
\toprule[1.25pt]
\multirow{2}{*}{\textbf{\shortstack{Guideline\\Development\\Datasets}}} & 
\multicolumn{1}{c}{\textbf{Bike-sharing}}  & \multicolumn{1}{c}{\textbf{Metro}} & \textbf{EV}\\[2pt]
\cmidrule(lr){2-2} \cmidrule(lr){3-3} \cmidrule(lr){4-4}
& \textit{Chicago} & \textit{Shanghai} & \textit{Beijing}\\[2pt]
\midrule
Time Span & 2013.07-2014.09 & 2016.07-2016.09 & 2018.03-2018.05 \\
\# Locations & 585 & 288 & 629\\
\# Weather States & 24 & 24 & 24\\
\# Holiday & 136 & 12 & 13\\
\# POIs Categories & 299 & 14 & 14\\
\# Roads & 17,020 & 24,781 & 33,788\\
Demographic & \multicolumn{3}{c}{Population, Education, Income}\\
\midrule[1.25pt]

\multirow{2}{*}{\textbf{\shortstack{Guideline\\Evaluation\\Datasets}}} & \multicolumn{1}{c}{\textbf{Bike-sharing}}  & \multicolumn{1}{c}{\textbf{Speed}} & \textbf{Pedestrian}\\[2pt]
\cmidrule(lr){2-2} \cmidrule(lr){3-3} \cmidrule(lr){4-4} 
& \textit{NYC} & \textit{Bay Area} & \textit{Melbourne}\\[2pt]
\midrule
Time Span & 2013.07.-2014.09 & 2017.01-2017.07 & 2021.01-2022.11 \\
\# Locations & 820 & 325 & 55\\
\# Holiday & 136 & 56 & 204\\
\bottomrule[1.25pt]
		\end{tabular}}
	\end{center}
\end{table}

\section{Implementation Details} \label{implementation_details}
To incorporate different temporal knowledge, following previous research \cite{STMeta}, the inputs of the \textit{STMGCN} \cite{geng2019spatiotemporal} and \textit{STMeta} \cite{STMeta} networks consist of 17 historical observations, including six closeness records, seven daily records, and four weekly records. For the considerations of spatial knowledge, distance, and correlation graphs are introduced. The distance graphs are calculated based on the Euclidean distance. The correlation graphs are computed by the Pearson coefficient of the time series of stations. In the distance graph, we link nodes whose distances are smaller than the given distance threshold. In the correlation graph, we link nodes whose Pearson coefficients are bigger than a given correlation threshold. The thresholds we use are in Table~\ref{graphThresholds}. The degree of the graph Laplacian is set to 1. The hidden states of \textit{STMeta} and \textit{STMGCN} backbone networks are both 64 (the dimension of spatiotemporal embedding). Our experiment platform is a server with 8 CPU cores (11th Gen Intel(R) Core(TM) i7-11700K @ 3.60GHz), 32 GB RAM, and one GPU (NVIDIA TITAN Xp). We use Python 3.6.5 with TensorFlow on Ubuntu Linux release 5.11.1 (Core).

\begin{table}[htbp]
	\small
	\caption{Spatial relationship graph thresholds}
    \vspace{-.5em}
	\label{graphThresholds}
	\begin{center}
\begin{tabular}{lccccccc}
\toprule
& \multicolumn{1}{c}{\textbf{Bike-sharing}}  & \multicolumn{1}{c}{\textbf{Metro}} & \textbf{EV}\\
\midrule
Distance Threshold & 1000 & 5000 & 1000 \\
Correlation Threshold & 0.0 & 0.35 & 0.1 \\ 
\bottomrule
	\end{tabular}
	\end{center}
 \vspace{-2em}
\end{table}

\section{Effect of the embedding output dimension} \label{embedding_size_analysis}
Embedding layers need to be carefully set since the output size may be too small to indicate the contextual situation or be overlarge, which may result in the overfitting issue. We tune the output dimensions of embedding layers through search (Figure~\ref{embedding_size}). The output dimensions of a single embedding layer (including \emph{Emb-Concat}, \emph{Emb-Add}, and \emph{Emb-Gating}) are set to 16. The output dimensions of multiple embedding layers (including \emph{MultiEmb-Concat}, \emph{MultiEmb-Add}, and \emph{MultiEmb-Gating}) are set to `8-1-8-8', corresponding to weather, holiday, temporal position, and POIs.

\begin{figure}[h]
\centering
\setcounter{figure}{12}
\subfloat[Performance vs. single embedding layer]{
\includegraphics[width=0.46\linewidth]{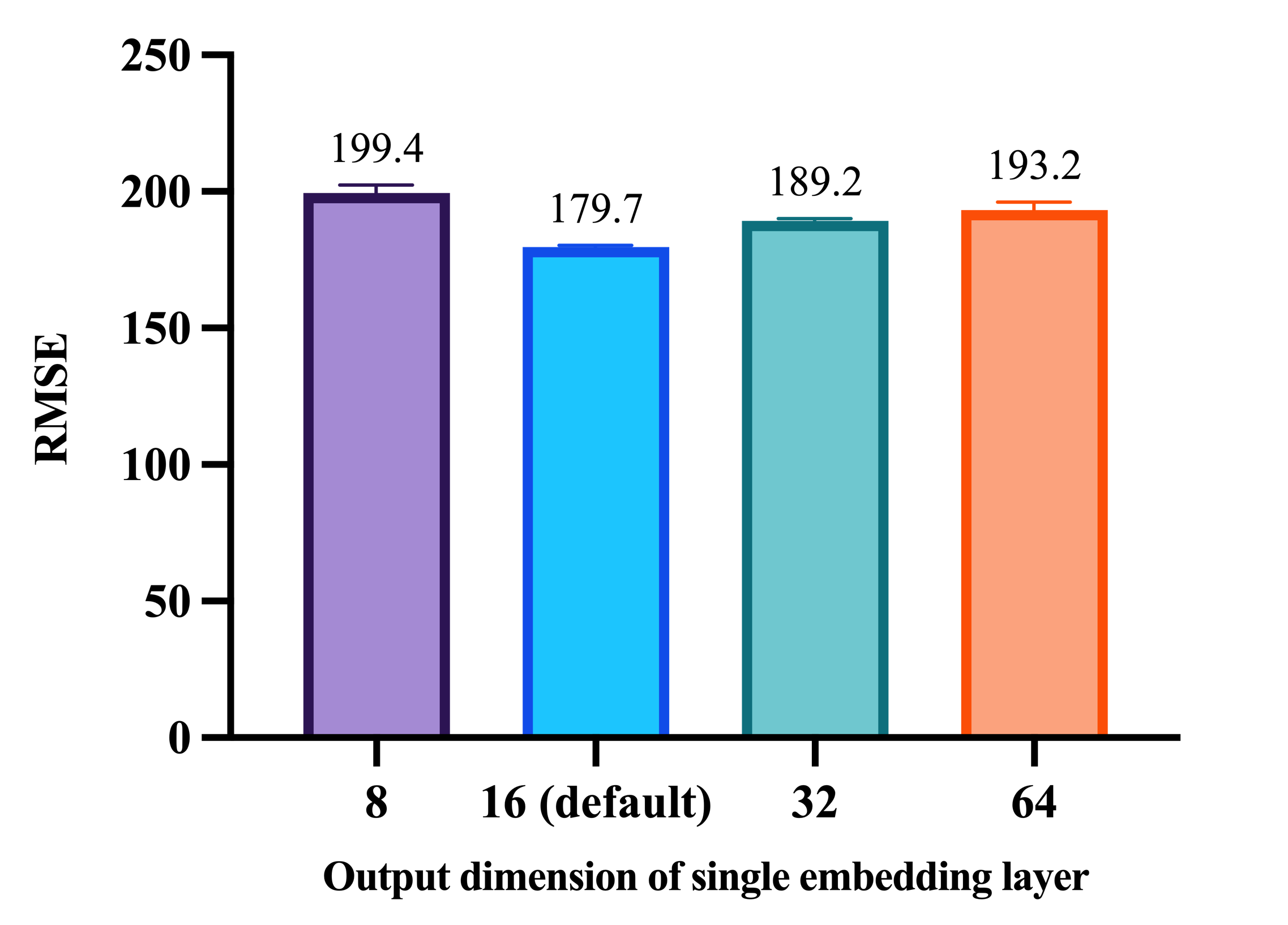}
\label{fig:acf_recall}
}
\hspace{1pt} 
\subfloat[Performance vs. multiple embedding layers]{
\includegraphics[width=0.46\linewidth]{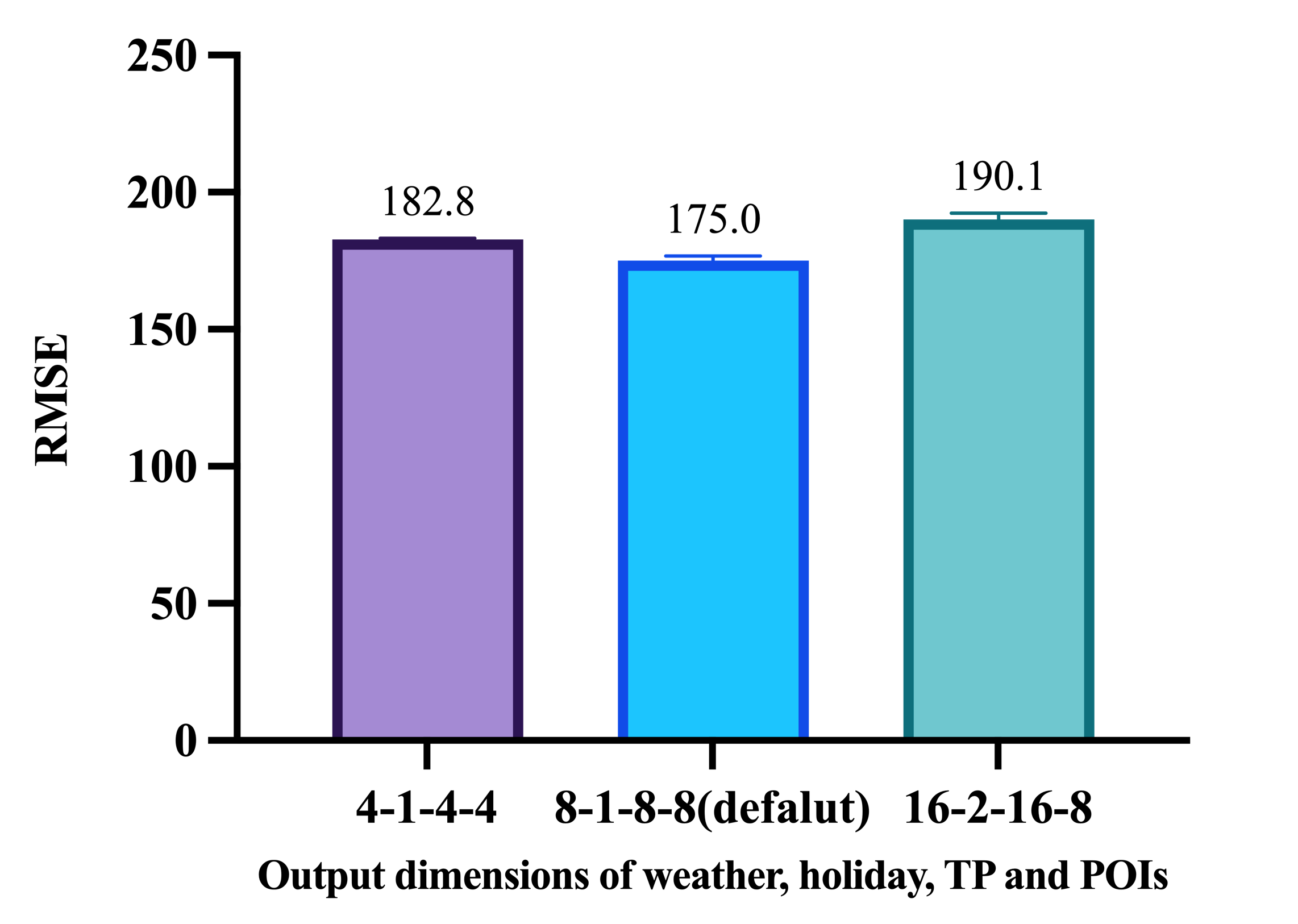}
\label{fig:scalability}
}
\caption{Effect of the output dimension of the single and the multiple embedding layers. These two experiments are based on \emph{Emb-Concat} and \emph{MultiEmb-Concat}, respectively.}
\label{embedding_size}
\end{figure}

\section{Data Analysis on Spatial Contextual Features} \label{spatial_data_analysis}

\begin{figure}[h]
	\centering
	\includegraphics[width=.9\linewidth]{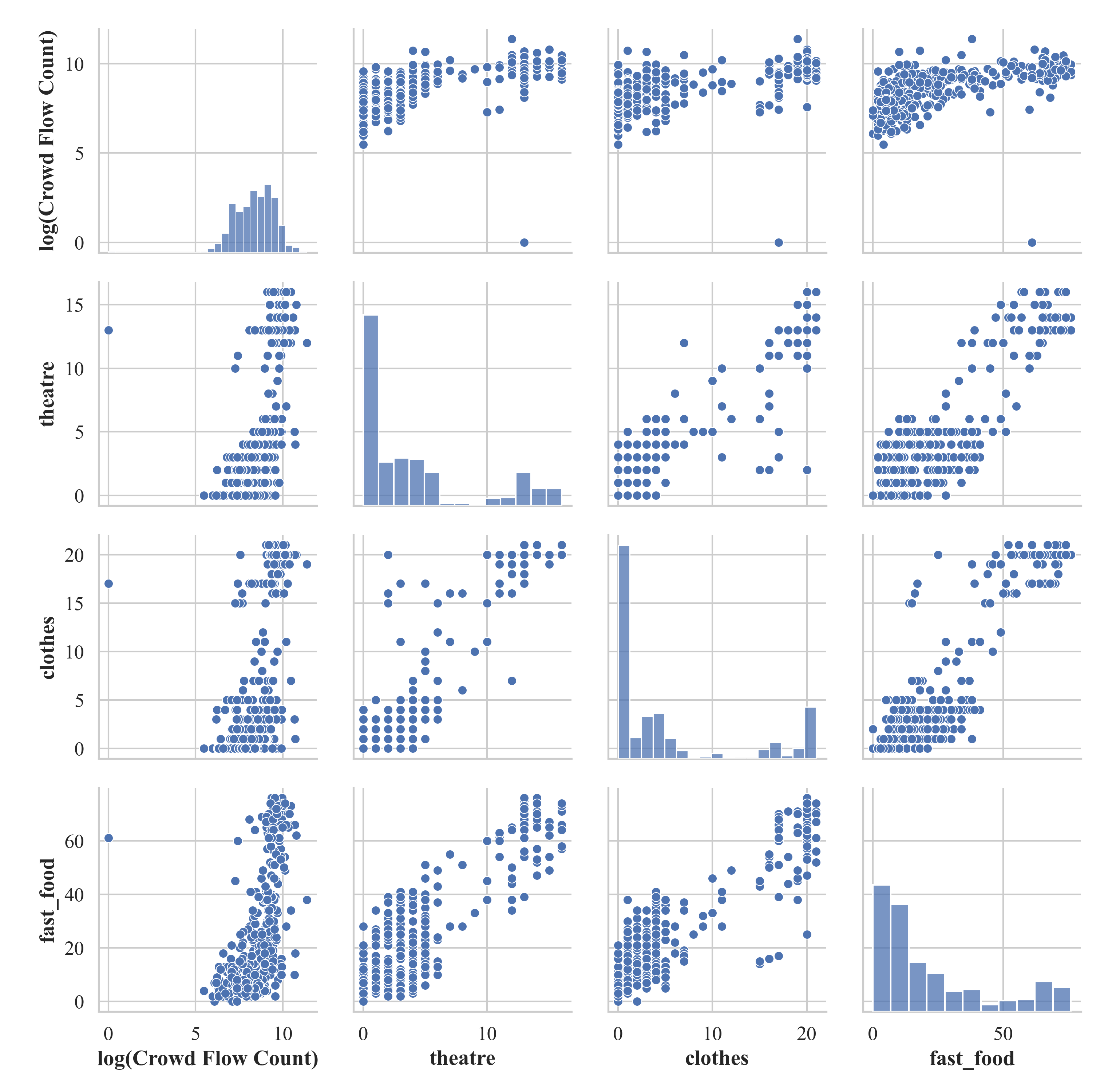}
    \vspace{-1em}
	\caption{Correlation matrix between POIs and crowd flow data. Diagonal elements represent univariate distributions, while off-diagonal elements depict correlations. We calculate the number of spatial features for each context type and the crowd flow at each location.}
	\label{fig: poi_plot}
\end{figure}

\begin{figure}[h]
	\centering
	\includegraphics[width=.90\linewidth]{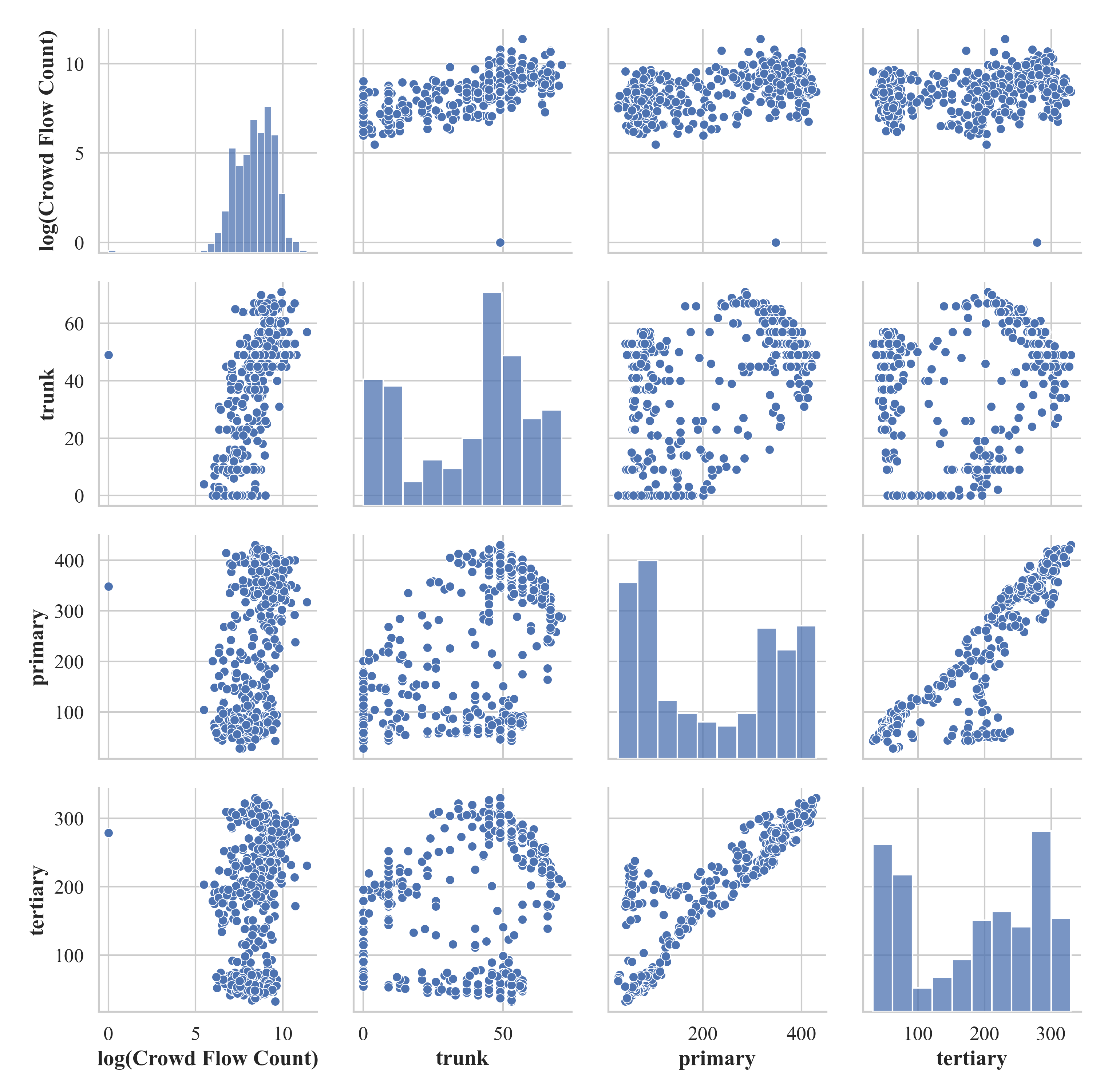}
    \vspace{-1em}
	\caption{Correlation matrix between road and crowd flow data. We show the three most correlated types of roads, with trunk roads showing the strongest correlation (Pearson coefficient = 0.5060).}
	\label{fig: road_plot}
\end{figure}

\begin{figure}[h]
	\centering
	\includegraphics[width=.90\linewidth]{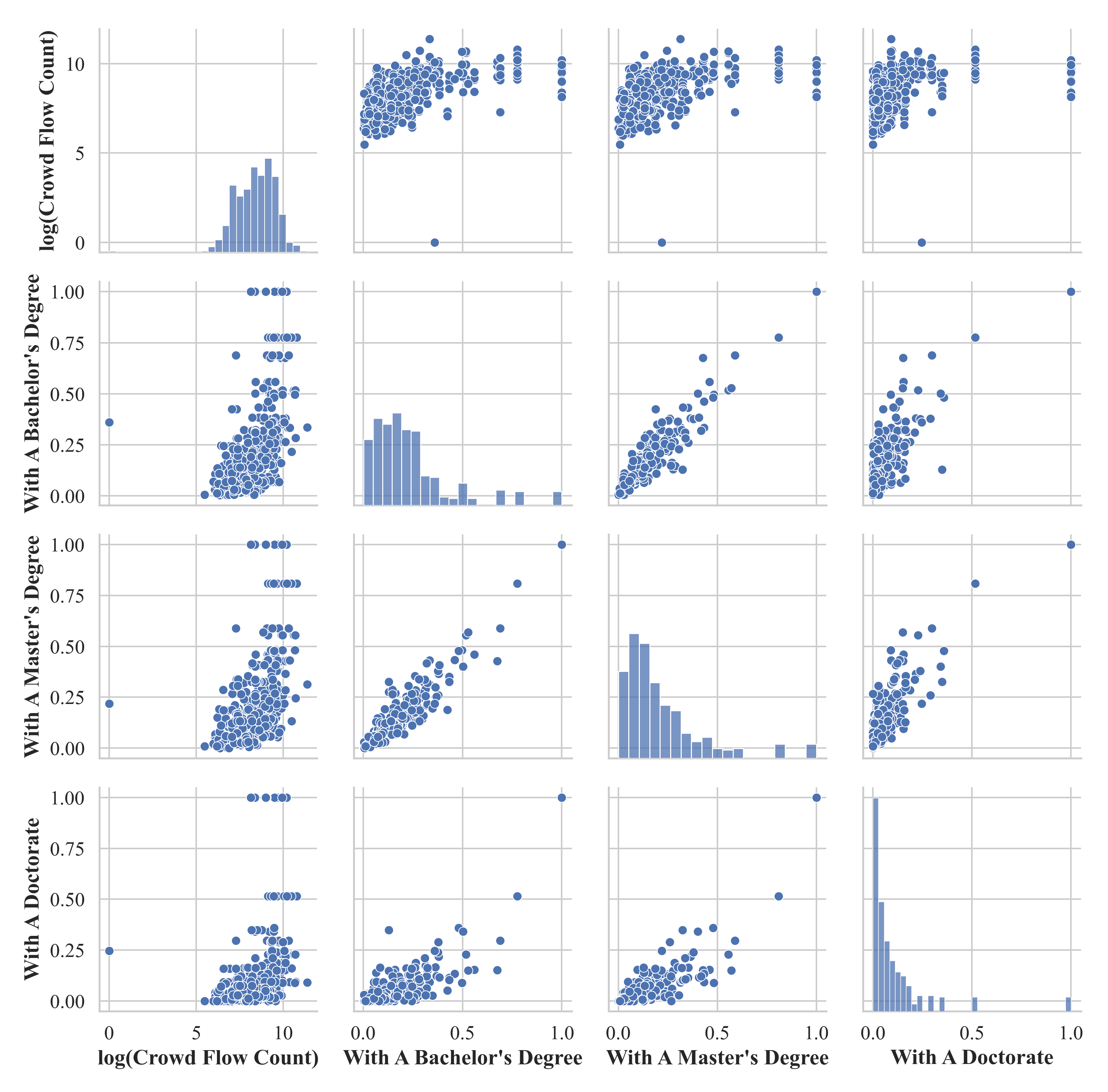}
    \vspace{-1em}
	\caption{Correlation matrix between demographic and crowd flow data. We show the three most correlated demographic features, with the population with a bachelor’s degree showing the strongest correlation (Pearson coefficient is 0.5007).}
	\label{fig: demo_plot}
\end{figure}

Figure~\ref{fig: poi_plot}, \ref{fig: road_plot}, and \ref{fig: demo_plot} show correlation matrices between different spatial contextual features (i.e., POIs, roads, demographics) and crowd flow data.
Specifically, we aggregate flows across all time slots to calculate the total volume for each location, reducing the crowd flow time series $\textbf{X} \in \mathbb{R}^{T \times N \times D}$ to $\mathbb{R}^{N \times D}$. Then, we retrieve nearby contextual features within a 5 km radius, following prior studies \cite{deep_fusion_net_2018}, and compute the number of spatial features for each context type.
We apply a log transformation to the crowd flow data to address scale differences across locations. 
The diagonal histograms show univariate distributions, while the off-diagonal plots reveal the correlation between variable pairs. Strong correlations are indicated by noticeable linear trends in the off-diagonal plots. For example, Figure~\ref{fig: poi_plot} shows a clear correlation (Pearson coefficient is 0.6390) between the number of theaters (a type of POI) and crowd flow.

\section{Additional MAE Results} \label{mae_results}
We here list the MAE results of different context modeling techniques (Table \ref{tab: techniques_60_STMeta_STMGCN_MAE} and Table \ref{tab: techniquues_30_120_STMeta_MAE}) and the MAE results of different combinations of contextual features (Table \ref{tab: features_60_STMeta_MAE} and Table \ref{tab: features_30_120_STMeta_MAE}).

\begin{table*}[htbp]
	\footnotesize
	\caption{60-minute MAE results of different modeling techniques based on \textit{STMeta} and \textit{STMGCN}. The best results are in bold. The modeling techniques marked with * significantly ($p < 0.05$) outperform \emph{No Context}. Those with significantly better \textit{avgNMAE} than \emph{No Context} are highlighted in {\color{blue!70}blue}.}
    \vspace{-2em}
	\label{tab: techniques_60_STMeta_STMGCN_MAE}
	\renewcommand\tabcolsep{4.0pt} 
	\begin{center}
    \begin{tabular}{lccccccccccccccccccccccccccccccccc}
	\toprule
	& \multicolumn{4}{c}{\textbf{STMeta}} & \multicolumn{4}{c}{\textbf{STMGCN}}\\
	\cmidrule(lr){2-5}  \cmidrule(lr){6-9} 
			
	\multicolumn{1}{c}{} & \multicolumn{1}{c}{\textit{Bike}} & \multicolumn{1}{c}{\textit{Metro}} & \multicolumn{1}{c}{\textit{EV}} & \multicolumn{1}{c}{\textit{avgNMAE}} & \multicolumn{1}{c}{\textit{Bike}} & \multicolumn{1}{c}{\textit{Metro}} & \multicolumn{1}{c}{\textit{EV}} & \multicolumn{1}{c}{\textit{avgNMAE}} \\
	\midrule
\textit{No Context} & 1.610$\pm$0.043 & 72.3$\pm$2.89 & 0.458$\pm$0.003 & 1.085$\pm$0.048 & 1.640$\pm$0.029 & 77.0$\pm$2.98 & 0.459$\pm$0.005 & 1.052$\pm$0.034 \\
\midrule
\multicolumn{2}{l}{\textbf{Early Joint Modeling}} \\
\textit{EarlyConcat} & 1.824$\pm$0.037 & 136.6$\pm$8.96 & 0.943$\pm$0.124 & 1.827$\pm$0.567 & 1.607$\pm$0.029 & 76.7$\pm$3.40 & 0.467$\pm$0.012 & 1.050$\pm$0.032 \\
\textit{EarlyAdd} & 1.631$\pm$0.099 & 102.3$\pm$29.2 & 0.472$\pm$0.036 & 1.239$\pm$0.273 & 1.677$\pm$0.047 & 78.9$\pm$5.42 & 0.476$\pm$0.018 & 1.081$\pm$0.030 \\
\midrule
\multicolumn{2}{l}{\textbf{Late Fusion}} \\
\textit{Raw-Concat} & 1.555$\pm$0.062 & 95.7$\pm$29.4 & 0.458$\pm$0.024 & 1.179$\pm$0.240 & 1.612$\pm$0.041 & 74.5$\pm$3.02 & 0.492$\pm$0.027 & 1.059$\pm$0.044 \\
\textit{Raw-Add} & 1.541*$\pm$0.044 & 78.4$\pm$12.3 & 0.490$\pm$0.023 & 1.116$\pm$0.102 & 1.644$\pm$0.035 & 79.4$\pm$1.55 & 0.476$\pm$0.006 & 1.077$\pm$0.041 \\
\textit{Raw-Gating} & 1.527*$\pm$0.062 & \textbf{68.3$\pm$9.60} & 0.444*$\pm$0.009 & \cellcolor{blue!10}\textbf{1.026*$\pm$0.024} & 1.589*$\pm$0.032 & 75.9$\pm$1.20 & \textbf{0.455$\pm$0.004} & \cellcolor{blue!10}\textbf{1.034*$\pm$0.040} \\
\textit{Emb-Concat} & 1.564$\pm$0.064 & 80.2$\pm$13.2 & 0.444*$\pm$0.007 & 1.094$\pm$0.115 & 1.636$\pm$0.034 & 74.2$\pm$3.51 & 0.461$\pm$0.011 & 1.040$\pm$0.032 \\
\textit{Emb-Add} & 1.549$\pm$0.032 & 80.9$\pm$15.8 & 0.444*$\pm$0.010 & 1.094$\pm$0.125 & 1.593$\pm$0.028 & 79.8$\pm$1.77 & 0.474$\pm$0.009 & 1.066$\pm$0.038 \\

\textit{Emb-Gating} & 1.536*$\pm$0.061 & 71.8$\pm$6.24 & 0.448*$\pm$0.012 & 1.051$\pm$0.086 & 1.580*$\pm$0.024 & 75.8$\pm$2.53 & 0.459$\pm$0.009 & \cellcolor{blue!10}\textbf{1.034*$\pm$0.031} \\
\textit{MultiEmb-Concat} & 1.534*$\pm$0.074 & 80.8$\pm$11.3 & 0.456$\pm$0.013 & 1.101$\pm$0.123 & 1.592$\pm$0.023 & \textbf{73.2$\pm$4.63} & 0.498$\pm$0.026 & 1.053$\pm$0.045 \\
\textit{MultiEmb-Add} & 1.589$\pm$0.085 & 84.5$\pm$17.8 & 0.441*$\pm$0.010 & 1.118$\pm$0.151 & 1.609$\pm$0.018 & 78.1$\pm$2.54 & 0.471$\pm$0.014 & 1.059$\pm$0.031 \\
\textit{MultiEmb-Gating} & 1.546$\pm$0.039 & 70.6$\pm$1.71 & 0.445*$\pm$0.012 & \cellcolor{blue!10}1.046*$\pm$0.047 & \textbf{1.547*$\pm$0.021} & 83.0$\pm$8.96 & 0.470$\pm$0.007 & 1.069$\pm$0.080 \\
\textit{LSTM-Concat} & 1.514*$\pm$0.050 & 70.9$\pm$5.01 & 0.444*$\pm$0.005 & \cellcolor{blue!10}1.038*$\pm$0.047 & 1.602$\pm$0.016 & 84.1$\pm$5.27 & 0.475$\pm$0.005 & 1.089$\pm$0.071 \\
\textit{LSTM-Add} & \textbf{1.507*$\pm$0.058} & 70.0$\pm$5.84 & \textbf{0.440*$\pm$0.004} & \cellcolor{blue!10}1.029*$\pm$0.042 & 1.603$\pm$0.032 & 81.1$\pm$4.36 & 0.474$\pm$0.011 & 1.074$\pm$0.049 \\
\textit{LSTM-Gating} & 1.510*$\pm$0.052 & 73.9$\pm$10.7 & 0.449*$\pm$0.012 & 1.055$\pm$0.069 & 1.566*$\pm$0.051 & 84.9$\pm$2.87 & 0.463$\pm$0.009 & 1.076$\pm$0.091 \\
	\bottomrule
	\end{tabular}
	\end{center}
\end{table*}

\begin{table}[htbp!]
	\footnotesize
	\caption{30/120-minute MAE results of different modeling techniques based on \textit{STMeta}. The best results are highlighted in bold. \emph{No Context} does not incorporate any contextual features. The modeling techniques with better \textit{avgNMAE} than \emph{No Context} are highlighted in {\color{blue!70}blue}.}
    \vspace{-2em}
	\label{tab: techniquues_30_120_STMeta_MAE}
    \renewcommand\tabcolsep{2pt}
	\begin{center}
	\resizebox{0.49\textwidth}{!}{
	
    \begin{tabular}{lccccccccccccccccccccccc}
	\toprule
	& \multicolumn{4}{c}{\textbf{30-minute}} & \multicolumn{4}{c}{\textbf{120-minute}}\\
	\cmidrule(lr){2-5} \cmidrule(lr){6-9}

    \multicolumn{1}{c}{} & \multicolumn{1}{c}{\textit{Bike}} & \multicolumn{1}{c}{\textit{Metro}} & \multicolumn{1}{c}{\textit{EV}} & \multicolumn{1}{c}{\textit{avgNMAE}} & \multicolumn{1}{c}{\textit{Bike}} & \multicolumn{1}{c}{\textit{Metro}} & \multicolumn{1}{c}{\textit{EV}} & \multicolumn{1}{c}{\textit{avgNMAE}} \\

\midrule

\textit{No Context} & 1.374 & 38.37 & 0.342 & 1.048 & 2.093 & 139.4 & 0.604 & 1.068 \\ 
\textit{EarlyConcat} & 1.961 & 64.95 & 0.683 & 1.783 & 2.521 & 227.7 & 0.701 & 1.417 \\ 
\textit{EarlyAdd} & 1.644 & 39.40 & 0.380 & 1.164 & 2.147 & 159.1 & 0.671 & 1.168 \\ 
\textit{Raw-Concat} & 1.382 & 46.95 & 0.340 & 1.125 & 2.081 & 356.2 & 0.594 & 1.591 \\ 
\textit{Raw-Add} & 1.417 & 47.27 & 0.366 & 1.163 & 2.053 & 370.3 & 0.645 & 1.653 \\ 

\textit{Raw-Gating} & 1.394 &  \textbf{36.51} & 0.338 &  \cellcolor{blue!10}\textbf{1.031} & 2.061 &  \textbf{136.0} & 0.558 &  \cellcolor{blue!10}\textbf{1.026} \\ 
\textit{Emb-Concat} & 1.372 & 41.32 & 0.345 & 1.076 & 2.063 & 171.5 & 0.554 & 1.110 \\

\textit{Emb-Add} & \textbf{1.295} & 41.85 & 0.341 & 1.057 & 2.053 & 142.4 & 0.556 & \cellcolor{blue!10}1.039 \\ 
\textit{Emb-Gating} & 1.349 & 48.65 & 0.347 & 1.140 & 2.066 & 175.1 & 0.552 & 1.119 \\ 
\textit{MultiEmb-Concat} & 1.350 & 42.27 & 0.342 & 1.076 & 2.062 & 145.9 & 0.590 & 1.070 \\ 
\textit{MultiEmb-Add} & 1.458 & 49.72 & 0.342 & 1.172 & 2.060 & 181.5 & 0.584 & 1.153 \\ 
\textit{MultiEmb-Gating} & 1.360 & 43.45 & 0.342 & 1.090 & 2.063 & 169.0 & \textbf{0.530} & 1.089 \\ 

\textit{LSTM-Concat} & 1.380 & 38.86 & 0.340 & 1.051 & 2.070 & 150.4 & 0.556 & \cellcolor{blue!10}1.061 \\  

\textit{LSTM-Add} & 1.334 & 38.92 &  \textbf{0.332} & \cellcolor{blue!10}1.032 & 2.064 & 141.7 & 0.551 & \cellcolor{blue!10}1.036 \\ 

\textit{LSTM-Gating} & 1.390 & 39.40 & 0.333 & 1.051 & \textbf{2.012} & 147.7 & 0.556 & \cellcolor{blue!10}1.045 \\ 
\bottomrule
	\end{tabular}}
	\end{center}
\end{table}

\begin{table}[htbp]
	\footnotesize
	\caption{60-minute MAE results of different contextual features based on \textit{STMeta}. The best results are in bold. The feature combinations with significantly ($p < 0.05$) better \textit{avgNMAE} than \emph{No Context} are in {\color{blue!70}blue}. (Wea: Weather; Holi: Holiday; TP: Temporal Position; POIs: Point of Interests; Demo: Demographic; SP: Spatial Position)}
    \vspace{-1em}
	\label{tab: features_60_STMeta_MAE}
    \renewcommand\tabcolsep{2.0pt} 
	\begin{center}
    \resizebox{0.48\textwidth}{!}{
    \begin{tabular}{lcccccccccccccccccccccccccccccccccc}
    \toprule
    \multicolumn{5}{c}{\textbf{STMeta}}\\
    \cmidrule(lr){2-4} \cmidrule(lr){5-5}
    & \textit{Bike} & \textit{Metro} & \textit{EV} & \textit{avgNMAE} \\
    \midrule

\textit{No Context} & 1.610$\pm$0.043 & 72.3$\pm$2.89 & 0.458$\pm$0.003 & 1.114$\pm$0.103  \\ 
\midrule
\multicolumn{2}{l}{\textbf{Temporal Contextual Feature}} \\
\textit{Wea} & 1.562$\pm$0.019 & 77.7$\pm$7.99 & 0.463$\pm$0.007 & 1.139$\pm$0.160 \\ 

\textit{Holi} & 1.553*$\pm$0.004 & 68.7$\pm$2.04 & 0.441*$\pm$0.003 & \cellcolor{blue!10}1.069*$\pm$0.088  \\ 

\textit{TP} & 1.522*$\pm$0.027 & 60.6*$\pm$1.26 & 0.451$\pm$0.005 & \cellcolor{blue!10}1.024*$\pm$0.017  \\ 

\textit{Wea-Holi} & 1.570$\pm$0.015 & 77.1$\pm$5.72 & 0.459$\pm$0.006 & 1.134$\pm$0.155 \\ 
\textit{Wea-TP} & \textbf{1.508*$\pm$0.033} & 79.8$\pm$16.4 & 0.451$\pm$0.008 & 1.129$\pm$0.199  \\ 

\textit{Holi-TP} & \textbf{1.508*$\pm$0.032} & 61.4*$\pm$0.94 & \textbf{0.440*$\pm$0.001} & \cellcolor{blue!10}\textbf{1.017*$\pm$0.025}  \\ 
\textit{Wea-Holi-TP} & 1.527$\pm$0.062 & 68.3$\pm$9.60 & 0.444*$\pm$0.009 & 1.102$\pm$0.142  \\ 
\midrule
\multicolumn{2}{l}{\textbf{Spatial Contextual Feature}} \\

\textit{POIs} & 1.589$\pm$0.008 & 67.8*$\pm$1.31 & 0.451$\pm$0.006 & \cellcolor{blue!10}1.079*$\pm$0.068  \\ 
\textit{Road} & 1.553*$\pm$0.009 & 71.2$\pm$4.79 & 0.448*$\pm$0.003 & 1.088$\pm$0.108 \\ 
\textit{Demo} & 1.556*$\pm$0.004 & 74.8$\pm$3.65 & 0.458$\pm$0.008 & 1.117$\pm$0.136 \\ 
\textit{SP} & 1.569$\pm$0.015 & 73.7$\pm$6.74 & 0.449*$\pm$0.006 & 1.107$\pm$0.129 \\ 

\midrule
\multicolumn{5}{l}{\textbf{Temporal Contextual Feature \& Effective Spatial Contextual Feature}} \\
\textit{Wea-POIs} & 1.585$\pm$0.013 & 81.7$\pm$3.97 & 0.465$\pm$0.005 & 1.168$\pm$0.193  \\ 
\textit{Holi-POIs} & 1.578$\pm$0.009 & 67.4*$\pm$0.65 & 0.444*$\pm$0.002 & 1.071$\pm$0.127   \\ 

\textit{TP-POIs} & 1.534*$\pm$0.005 & \textbf{60.5*$\pm$1.62} & 0.447*$\pm$0.003 & \cellcolor{blue!10}1.022*$\pm$0.016  \\ 

\textit{Wea-Holi-POIs} & 1.567$\pm$0.012 & 80.6$\pm$7.28 & 0.455$\pm$0.006 & 1.150$\pm$0.192  \\ 
\textit{Wea-TP-POIs} & 1.533*$\pm$0.008 & 76.7$\pm$7.05 & 0.456$\pm$0.002 & 1.122$\pm$0.161 \\ 

\textit{Holi-TP-POIs} & 1.537*$\pm$0.005 & 62.8*$\pm$3.16 & 0.446*$\pm$0.002 & \cellcolor{blue!10}1.036*$\pm$0.034  \\ 
\textit{Wea-Holi-TP-POIs} & 1.533*$\pm$0.008 & 72.3$\pm$1.10 & 0.454$\pm$0.002 & 1.095$\pm$0.118 \\  

\midrule
\multicolumn{5}{l}{\textbf{Spatial Contextual Feature \& Effective Temporal Contextual Feature}} \\

\textit{Holi-TP-POIs} & 1.537*$\pm$0.005 & 62.8*$\pm$3.16 & 0.446*$\pm$0.003 & \cellcolor{blue!10}1.036*$\pm$0.034 \\ 

\textit{Holi-TP-Road} & 1.523*$\pm$0.003 & 67.8*$\pm$2.00 & 0.447*$\pm$0.002 & \cellcolor{blue!10}1.062*$\pm$0.080 \\ 

\textit{Holi-TP-Demo} & 1.528*$\pm$0.011 & 68.8$\pm$2.39 & 0.441*$\pm$0.002 & \cellcolor{blue!10}1.064*$\pm$0.094 \\ 

\textit{Holi-TP-SP} & 1.533*$\pm$0.013 & 69.8$\pm$2.28 & 0.444*$\pm$0.002 & \cellcolor{blue!10}1.073*$\pm$0.100 \\ 
\midrule
\textit{All} & 1.546*$\pm$0.003 & 79.9$\pm$3.08 & 0.446*$\pm$0.003 & 1.135$\pm$0.196 \\  
\bottomrule
\end{tabular}}
	\end{center}
\end{table}

\clearpage

\begin{table}[htbp]
	\footnotesize
	\caption{30/120-minute MAE results of different contextual features based on \textit{STMeta}. The best results are in bold. The feature combinations with better \textit{avgNMAE} than \emph{No Context} are highlighted in {\color{blue!70}blue}.}
    \vspace{-1.5em}
	\label{tab: features_30_120_STMeta_MAE}
	\renewcommand\tabcolsep{2pt}
	\begin{center}
    \resizebox{0.48\textwidth}{!}{
	\begin{tabular}{lccccccccccccccccccccccccccc}
\toprule

\multirow{2}{*}{\textbf{STMeta}} & \multicolumn{4}{c}{\textbf{30-minute}} & \multicolumn{4}{c}{\textbf{120-minute}} \\

\cmidrule(lr){2-5} \cmidrule(lr){6-9} 

& Bike & Metro & EV & \textit{avgNMAE} & Bike & Metro & EV & \textit{avgNMAE} \\
\midrule
\textit{No Context} & 1.374 & 38.37 & 0.342 & 1.061 & 2.093 & 139.4 & 0.604 & 1.144 \\ 
\midrule
\textit{Wea} & 1.452 & 41.99 & 0.337 & 1.109 & 2.072 & 189.9 & 0.596 & 1.277 \\ 

\textit{Holi} & 1.359 & 37.54 & 0.331 & \cellcolor{blue!10}1.038 & 1.903 & 138.4 & 0.551 & \cellcolor{blue!10}1.074 \\ 

\textit{TP} & \textbf{1.318} & 35.98 & 0.325 & \cellcolor{blue!10}\textbf{1.006} & 1.922 & 122.1 & 0.571 & \cellcolor{blue!10}1.045 \\ 
\textit{Wea-Holi} & 1.467 & 40.50 & 0.339 & 1.101 & 2.068 & 148.4 & 0.606 & 1.167 \\ 

\textit{Wea-TP} & 1.393 & 39.62 & 0.336 & 1.071 & 1.969 & 133.0 & 0.579 & \cellcolor{blue!10}1.089 \\ 

\textit{Holi-TP} & 1.344 & 36.62 & \textbf{0.324} & \cellcolor{blue!10}1.018 & \textbf{1.898} & \textbf{119.0} & \textbf{0.521} & \cellcolor{blue!10}\textbf{1.000}\\ 

\textit{Wea-Holi-TP} & 1.394 & 39.51 & 0.328 & 1.062 & 2.021 & 136.0 & 0.558 & \cellcolor{blue!10}1.093 \\ 
\midrule

\textit{POIs} & 1.361 & 38.10 & 0.335 & \cellcolor{blue!10}1.048 & 2.158 & 133.6 & 0.592 & \cellcolor{blue!10}1.132\\
\textit{Road} & 1.357 & 41.25 & 0.330 & 1.071 & 2.110 & 136.1 & 0.643 & 1.163\\
\textit{Demo} & 1.340 & 38.95 & 0.339 & \cellcolor{blue!10}1.054 & 2.114 & 137.5 & 0.607 & 1.145\\
\textit{SP} & 1.393 & 38.81 & 0.336 & 1.063 & 2.108 & 144.1 & 0.596 & 1.155\\
\midrule
\textit{Wea-POIs} & 1.381 & 50.58 & 0.369 & 1.205 & 2.084 & 157.9 & 0.611 & 1.199 \\ 
\rowcolor{blue!10} 
\textit{Holi-POIs} & 1.361 & 37.72 & 0.344 & 1.053 & 2.145 & 142.8 & 0.564 & 1.138 \\ 

\textit{TP-POIs} & 1.333 & \textbf{35.41} & 0.344 & \cellcolor{blue!10}1.024 & 2.128 & 137.9 & 0.594 & \cellcolor{blue!10}1.140 \\ 
\textit{Wea-Holi-POIs} & 1.368 & 42.70 & 0.344 & 1.102 & 2.066 & 148.1 & 0.612 & 1.169 \\ 
\textit{Wea-TP-POIs} & 1.385 & 42.71 & 0.342 & 1.104 & 2.148 & 145.1 & 0.581 & 1.155 \\ 

\textit{Holi-TP-POIs} & 1.361 & 38.45 & 0.325 & \cellcolor{blue!10}1.041 & 2.139 & 120.9 & 0.571 & \cellcolor{blue!10}1.080 \\ 

\textit{Wea-Holi-TP-POIs} & 1.365 & 41.14 & 0.344 & 1.086 & 2.146 & 139.2 & 0.574 & \cellcolor{blue!10}1.134 \\ 
\midrule

\textit{Holi-TP-POIs} & 1.361 & 38.45 & 0.325 & \cellcolor{blue!10}1.041 & 2.139 & 120.9 & 0.571 & \cellcolor{blue!10}1.080 \\ 

\textit{Holi-TP-Road} & 1.355 & 39.91 & 0.336 & 1.064 & 2.047 & 136.5 & 0.563 & \cellcolor{blue!10}1.102\\ 

\textit{Holi-TP-Demo} & 1.359 & 35.52 & 0.332 & \cellcolor{blue!10}1.020 & 2.059 & 135.0 & 0.556 & \cellcolor{blue!10}1.096\\

\textit{Holi-TP-SP} & 1.352 & 35.85 & 0.327 & \cellcolor{blue!10}1.016 & 2.070 & 130.6 & 0.562 & \cellcolor{blue!10}1.089 \\  
\midrule

\textit{All} & 1.366 & 41.45 & 0.335 & 1.080 & 2.047 & 133.9 & 0.568 & \cellcolor{blue!10}1.098\\
\bottomrule
		\end{tabular}}
	\end{center}
\end{table}

\vspace{-2em}

\section{Results of Feature Combinations Based on XGBoost} \label{xgboost_results}
We here list the results of different contextual features based on \textit{XGBoost} in Table \ref{tab: features_60_XGBoost_RMSE}.

\begin{table}[h]
	\footnotesize
	\caption{60-minute RMSE results of different contextual features based on \textit{XGBoost}. The best results are in bold. The feature combinations with better \textit{avgNRMSE} than \emph{No Context} are in {\color{blue!70}blue}.}
	\label{tab: features_60_XGBoost_RMSE}
    \vspace{-1.5em}
	\begin{center}
    \resizebox{0.48\textwidth}{!}{
	\setlength{\tabcolsep}{2.0pt}{
    \begin{tabular}{lcccccccccccccccccccccccccccccccccc}
    \toprule
    
    & \multicolumn{4}{c}{\textbf{XGBoost}} \\
    
    \cmidrule(lr){2-5} 
& Bike & Metro & EV & \textit{avgNRMSE}  \\
\midrule
\textit{No Context} & 3.011$\pm$0.008 & 185.7$\pm$0.110 & 0.835$\pm$0.003 & 1.030$\pm$0.023  \\ 
\textit{Wea} & 2.970$\pm$0.005 & 188.5$\pm$0.010 & 0.848$\pm$0.004 & 1.036$\pm$0.031  \\ 

\textit{Holi} & 2.985$\pm$0.008 & 185.5$\pm$0.060 & 0.827$\pm$0.003 & \cellcolor{blue!10}1.024$\pm$0.019  \\ 

\textit{TP} & 2.955$\pm$0.008 & \textbf{184.3$\pm$0.170} & 0.795$\pm$0.006 & \cellcolor{blue!10}1.004$\pm$0.005  \\ 
\textit{Wea-Holi} & 2.952$\pm$0.006 & 188.5$\pm$0.150 & 0.840$\pm$0.002 & 1.031$\pm$0.027  \\ 

\textit{Wea-TP} & 2.934$\pm$0.006 & 188.0$\pm$0.130 & 0.808$\pm$0.003 & \cellcolor{blue!10}1.014$\pm$0.010  \\ 

\textit{Holi-TP} & 2.936$\pm$0.006 & 185.0$\pm$0.120 & \textbf{0.792$\pm$0.005} & \cellcolor{blue!10}\textbf{1.002$\pm$0.002}  \\ 

\textit{Wea-Holi-TP} & \textbf{2.927$\pm$0.005} & 187.6$\pm$0.030 & 0.805$\pm$0.003 & \cellcolor{blue!10}1.012$\pm$0.010  \\ 
\textit{POIs} & 3.011$\pm$0.008 & 185.7$\pm$0.110 & 0.835$\pm$0.003 & 1.030$\pm$0.023  \\ 
\textit{Wea-POIs} & 2.970$\pm$0.005 & 188.5$\pm$0.010 & 0.848$\pm$0.004 & 1.036$\pm$0.031  \\ 

\textit{Holi-POIs} & 2.985$\pm$0.008 & 185.5$\pm$0.060 & 0.827$\pm$0.003 & \cellcolor{blue!10}1.024$\pm$0.019  \\ 

\textit{TP-POIs} & 2.955$\pm$0.008 & \textbf{184.3$\pm$0.170} & 0.795$\pm$0.006 & \cellcolor{blue!10}1.004$\pm$0.005  \\ 
\textit{Wea-Holi-POIs} & 2.952$\pm$0.006 & 188.5$\pm$0.150 & 0.840$\pm$0.002 & 1.031$\pm$0.027  \\ 

\textit{Wea-TP-POIs} & 2.934$\pm$0.006 & 188.0$\pm$0.130 & 0.808$\pm$0.003 & \cellcolor{blue!10}1.014$\pm$0.010  \\ 

\textit{Holi-TP-POIs} & 2.936$\pm$0.006 & 185.0$\pm$0.120 & \textbf{0.792$\pm$0.007} & \cellcolor{blue!10}\textbf{1.002$\pm$0.003}  \\ 

\textit{Wea-Holi-TP-POIs} & \textbf{2.927$\pm$0.005} & 187.6$\pm$0.030 & 0.805$\pm$0.003 & \cellcolor{blue!10}1.012$\pm$0.010 \\ 
\bottomrule
    \end{tabular}}}
	\end{center}
\end{table}

\end{document}